\documentclass[10pt,twocolumn,letterpaper]{article}

\usepackage{cvpr}              % To produce the CAMERA-READY version
\usepackage[ruled,vlined,linesnumbered]{algorithm2e}
\usepackage{setspace}
\usepackage[accsupp]{axessibility}
\usepackage{pifont}
\usepackage{array}
\usepackage{array}
\usepackage{colortbl}
\usepackage{siunitx}
\usepackage{multirow}
\usepackage{graphicx}
\usepackage{amsmath}
\usepackage{rotating}
\usepackage{makecell}
\usepackage{float}
\usepackage{pifont}
\definecolor{cvprblue}{rgb}{0.21,0.49,0.74}
\usepackage[pagebackref,breaklinks,colorlinks,allcolors=cvprblue]{hyperref}

\def\paperID{} % *** Enter the Paper ID here
\def\confName{CVPR}
\def\confYear{2026}

\title{EnergyAction: Unimanual to Bimanual Composition with Energy-Based Models
}
\author{
Mingchen Song$^{1\,2}$,
Xiang Deng$^{1\,3\,\dagger}$,
Jie Wei$^{1}$, Dongmei Jiang$^{2}$, Liqiang Nie$^{1}$,Weili Guan$^{1\,3\,\dagger}$
\\ 
    $^1$Harbin Institute of Technology (Shenzhen),  \quad $^2$PengCheng Laboratory,
      \quad $^3$Shenzhen Loop Area Institute\\
      [1mm]
 \href{https://github.com/codeshop715/EnergyAction/}{https://github.com/codeshop715/EnergyAction}
  }

\begin{document}
\maketitle
\begin{abstract}
Recent advances in unimanual manipulation policies have achieved remarkable success across diverse robotic tasks through abundant training data and well-established model architectures. However, extending these capabilities to bimanual manipulation remains challenging due to the lack of bimanual demonstration data and the complexity of coordinating dual-arm actions. Existing approaches either rely on extensive bimanual datasets or fail to effectively leverage pre-trained unimanual policies. To address this limitation, we propose \textbf{EnergyAction}, a novel framework that compositionally transfers unimanual manipulation policies to bimanual tasks through the Energy-Based Models (EBMs). Specifically, our method incorporates three key innovations. First, we model individual unimanual policies as EBMs and leverage their compositional properties to compose left and right arm actions, enabling the fusion of unimanual policies into a bimanual policy. Second, we introduce an energy-based temporal-spatial coordination mechanism through energy constraints, ensuring the generated bimanual actions are both temporal coherence and spatial feasibility.
Third, we propose two different energy-aware denoising strategies that dynamically adapt denoising steps based on action quality assessment. These strategies ensure the generation of high-quality actions while maintaining superior computational efficiency compared to fixed-step denoising approaches. Experimental results demonstrate that EnergyAction effectively transfers unimanual knowledge to bimanual tasks, achieving superior performance on both simulated and real-world tasks with minimal bimanual data.
\end{abstract}
{
\let\thefootnote \relax \footnote{
$^\dagger$Corresponding author}
}
\section{Introduction}
\label{sec:intro}
%%%%%%%%%%%%%%%%%%%%%%%%%%%%%%%%%%%%%%%%%%%%%%%%%%%%%%%%%%%%%%%%%%%%%%%%%%%%%%%%
%introduciton部分第一段首先写双臂任务的应用场景，对比与单臂任务的区别（单臂相对双臂来说更多的数据更多的模型，较为成功）。同时指出双臂任务存在的问题：数据难以收集且相对于单臂的动作空间维度更大，需要考虑双臂的协同。
%第二段写现有的双臂方法有哪些，点出各类方法存在的问题。同时引出我们的动机：如何只利用少量的双臂数据，将单臂的policy迁移到双臂任务。
%第三段从能量模型切入，把双臂action生成问题转变成单臂能量模型的组合问题，提出一个优雅的解决方案：通过能量模型对单臂policy进行建模，利用其组合特性组合成双臂policy。
%第四段具体的描述我们的方法，分三点：1. an energy-based compositional architecture 2. an energy-based temporal-spatial coordination mechanism 3. an energy-aware denoising strategy。
%最后是贡献部分,总共是四点贡献。
%%%%%%%%%%%%%%%%%%%%%%%%%%%%%%%%%%%%%%%%%%%%%%%%%%%%%%%%%%%%%%%%%%%%%%%%%%%%%%%%
\begin{figure}[t!]
	\centering  
	\includegraphics[width=0.432\textwidth]{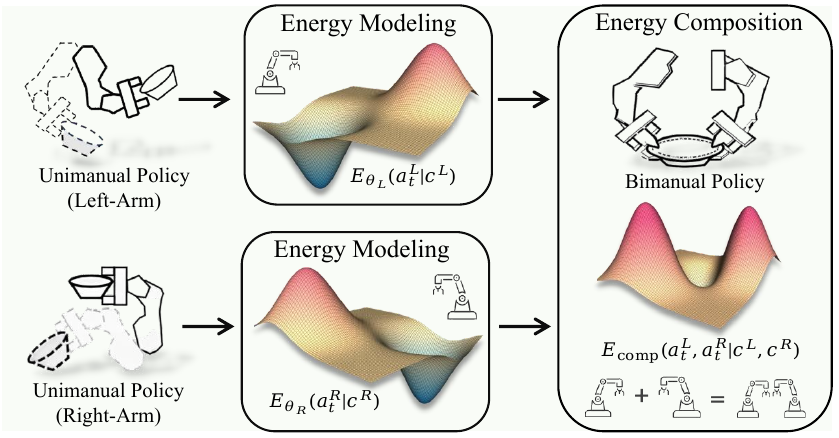}  
	\caption{\textbf{Compositional Action Generation.} Inspired by the theory of EBMs~\cite{hinton2002training}, we first model unimanual policies as energy functions and then leverage their compositional properties to obtain bimanual policies.} 
	\label{fig:1}  
\end{figure}  
\begingroup
\setlength{\leftmargini}{0pt}
\setlength{\rightmargin}{0pt}
\begin{quote} 
\itshape\small The art of science is to resolve difficulties by dividing them into as many simple parts as possible. \\
\mbox{}\hfill -- René Descartes (1596--1650)~\cite{descartes1912discourse}
\end{quote}
\endgroup
Bimanual manipulation~\cite{yang2025gripper,liu2024rdt} represents a fundamental capability for robotic systems to perform complex tasks that are difficult or impossible for single-arm robots, such as manipulating large objects, executing precise assembly operations, or performing collaborative motions. These capabilities make bimanual robots essential for applications ranging from household service~\cite{cao2024smart,zhang2024multi} to industrial assembly~\cite{buhl2019dual} and surgical assistance~\cite{hu2023towards}. Compared to unimanual systems, which have achieved impressive generalization through large-scale demonstration datasets~\cite{li2025learning} and scalable architectures~\cite{yan2025robotron}, bimanual manipulation faces significantly greater challenges due to the high-dimensional action space and the need for precise spatial-temporal coordination~\cite{chen2025learning,sun2024real}. Moreover, collecting high-quality bimanual demonstrations is prohibitively expensive, requiring sophisticated teleoperation systems and substantially higher human effort~\cite{darvish2023teleoperation,zhao2024aloha}, severely limiting the development of generalizable bimanual policies.

Current bimanual manipulation approaches typically design policies that directly model the joint action space of both arms using complex network architectures~\cite{kou2025roboannotatorx,li2026global}. However, the exponentially larger action space and the need to simultaneously learn individual manipulation skills, temporal synchronization~\cite{tomaselli2024synchronization}, and spatial compatibility~\cite{pan2025omnimanip,song2025robospatial} substantially increase the complexity of action generation and optimization difficulty. Existing methods struggle with coordination failures such as arm collisions and asynchronous movements~\cite{zhang2022matching,stoop2024method}, as they must implicitly learn physical constraints~\cite{zhong2025dexgrasp,zhou2025physvlm} from limited demonstrations~\cite{li2025sd2actor}. Some approaches simplify this by assigning fixed roles to each arm~\cite{varley2024embodied,gao2024bi}, but fail to generalize across tasks requiring flexible coordination. Recent efforts to learn robotic foundation models from large-scale bimanual data~\cite{zhang2025vlabench,wen2024diffusion} remain limited by prohibitive data collection costs. Motivated by these limitations, given the scarcity of bimanual data and the abundance of pre-trained unimanual policies with rich manipulation knowledge, a straightforward and intuitive idea emerges: \emph{Can we effectively transfer the rich manipulation knowledge embedded in pre-trained unimanual policies to bimanual tasks with minimal bimanual demonstrations?}

To address this challenge, we draw inspiration from the theory of EBMs~\cite{hinton2002training}, a classical approach with a long history across various domains that remains largely overlooked in modern robotic research. The key insight is that EBMs naturally support compositional generation through energy summation~\cite{du2020compositional,du2023reduce}, where complex concepts can be composed from simpler ones. This compositional property offers an elegant solution to our transfer learning problem: we decompose the complex bimanual action generation task into the composition of two unimanual policies. Building on these insights, we propose \textbf{EnergyAction}, a novel framework that compositionally transfers pre-trained unimanual policies to bimanual tasks. As illustrated in Figure~\ref{fig:1}, we first model unimanual policies as energy functions and then compose them through energy-based composition to generate coordinated bimanual actions.

% Inspired by the theory of Energy-Based Models (EBMs)~\cite{hinton2002training,du2020compositional,du2023reduce,xu2024energy}, where complex distributions can be generated through energy-based composition of different simple distributions.
% We propose \textbf{EnergyAction}, a novel framework that compositionally transfers pre-trained unimanual policies to bimanual tasks through EBMs. As shown in figure~\ref{fig:1}, 
%  we formulate individual unimanual policies as energy functions and compose them to generate coordinated bimanual actions, where left and right arm policies are treated as separate energy distributions combined through energy summation. 
% % This compositional approach enables effective knowledge transfer from pre-trained unimanual policies. Meanwhile, additional energy-based mechanisms are introduced to ensure the generated actions satisfy both coordination requirements and physical constraints. Furthermore, leveraging the energy landscape of EBMs, we develop an adaptive refinement strategy that improves inference efficiency while maintaining high-quality action generation.

Specifically, EnergyAction introduces three key innovations to enable effective knowledge transfer from unimanual to bimanual policies with minimal bimanual data. First, we develop an energy-based compositional architecture that models left and right arm policies as separate energy functions and composes them to generate bimanual actions. Second, to ensure coordination of generated bimanual actions, we introduce an energy-based temporal-spatial coordination mechanism that ensures generated actions satisfy both temporal coherence (via smoothness and synchronization constraints) and spatial feasibility (via collision avoidance in end-effector and joint space through inverse kinematics~\cite{zhao2024inverse}). Third, to improve inference efficiency while maintaining action quality, we propose two different energy-aware denoising strategies that dynamically adjust denoising steps based on action quality assessment. These strategies perform more denoising steps for high-energy configurations that indicate uncertainty or constraint violations, while efficiently handling low-energy cases with minimal steps. Compared to conventional diffusion policies~\cite{zhang2025flowpolicy,ma2024hierarchical,ke20243d} 
that apply fixed denoising steps, our method ensures high-quality action generation while maintaining superior inference efficiency.\\
% Second, we introduce an energy-based temporal-spatial coordination mechanism 
% that ensures generated actions satisfy both temporal coherence (via smoothness and synchronization constraints) and spatial feasibility (via collision avoidance in end-effector and joint space through inverse kinematics~\cite{zhao2024inverse}), with adaptive constraint weighting based on the current state. Third, we propose two different energy-aware denoising strategies that dynamically adjust denoising steps based on energy assessment. These strategies perform more denoising steps for high-energy configurations, which indicate uncertainty or constraint violations, while efficiently handling low-energy cases. Compared to conventional diffusion policies~\cite{zhang2025flowpolicy,ma2024hierarchical,ke20243d} that apply fixed denoising steps, our method ensures high-quality action generation while improving inference efficiency. \\
Our main contributions can be summarized as follows:
\begin{itemize}
    \item We propose an energy-based compositional architecture that models unimanual policies as energy functions and composes them through energy summation to generate bimanual actions, providing a principled and data-efficient approach to leverage existing unimanual policies for bimanual tasks learning.
    
    \item We introduce an energy-based temporal-spatial coordination mechanism that explicitly enforces temporal coherence and collision avoidance, ensuring generated bimanual actions satisfy both kinematic constraints and physical feasibility requirements.
    
    \item We develop two different energy-aware denoising strategies that dynamically adjust denoising steps based on action quality assessment, ensuring high-quality action generation while achieving superior inference efficiency compared to fixed-step denoising approaches.
    
    \item Through extensive experiments on simulated and real-world tasks, we demonstrate that EnergyAction achieves superior bimanual manipulation performance with minimal bimanual demonstrations, validating the effectiveness of our compositional transfer learning paradigm.
\end{itemize}

% \section{Related Work}
\section{Related Work}
%%%%%%%%%%%%%%%%%%%%%%%%%%%%%%%%%%%%%%%%%%%%%%%%%%%%%%%%%%%%%%%%%%%%%%%%%%%%%%%%
%Related Work包括两部分
%第一部分是从Robotic Manipulation这个任务切入，首先简单的介绍下单臂policy的成功，然后重点介绍下现有的双臂模型存在的问题和局限。最后引出我们的EnergyAction与其他方法的不同。
%第二部分是基于能量的模型。首先简单介绍下能量模型的基本形式和他的组合特性，然后说现有的能量模型应用的场景，包括语言，视觉生成等任务。最后引出我们的EnergyAction是第一个进行基于能量的单臂组合成双臂的方法。
%%%%%%%%%%%%%%%%%%%%%%%%%%%%%%%%%%%%%%%%%%%%%%%%%%%%%%%%%%%%%%%%%%%%%%%%%%%%%%%%
\subsection{Robotic Manipulation} 
Robotic manipulation~\cite{mete2407quest,yao2025think,song2025few} research has achieved remarkable progress  through carefully designed model architectures~\cite{ryu2024diffusion,li2025object,jia2025lift3d} and large-scale demonstration datasets~\cite{ye2024latent,su2025robosense,qi2025two}. For instance, methods such as the RT series of works~\cite{brohan2022rt,zitkovich2023rt} and the Octo~\cite{team2024octo} demonstrate strong generalization capabilities across diverse objects and tasks. However, bimanual manipulation poses substantial challenges including higher-dimensional action spaces, precise spatiotemporal coordination requirements~\cite{yang2024spatiotemporal,zhao2025dual}, and prohibitively expensive data collection~\cite{wang2024dexcap,mu2025robotwin}. To address these challenges, recent research has explored various innovative approaches. For example, 3DFA~\cite{gkanatsios20253d} integrates flow matching with 3D scene representations to enable efficient trajectory prediction, achieving fast training and state-of-the-art results on bimanual benchmarks. Similarly, AnyBimanual~\cite{lu2025anybimanual} attempts to leverage pre-trained unimanual policies through skill scheduling and visual alignment techniques to reduce the dependency on bimanual data. In contrast, our EnergyAction models unimanual policies as composable energy functions for bimanual action generation, ensuring robust temporal and spatial coordination through explicit energy-based constraints while enabling efficient inference via two different adaptive denoising processes.

\subsection{Energy-Based Models}

EBMs model probability distributions by assigning energy values to configurations~\cite{lecun2006tutorial}. For a state $x$, an EBM defines the distribution as $p_\theta(x) = \frac{\exp(-E_\theta(x))}{Z(\theta)}$, where $E_\theta(x)$ is the energy function and $Z(\theta) = \int \exp(-E_\theta(x))dx$ is the partition function. A key property of EBMs is compositionality: given energy functions $E_{\theta_1}(x)$ and $E_{\theta_2}(x)$, their joint distribution is $p(x) \propto \exp(-(E_{\theta_1}(x) + E_{\theta_2}(x)))$. This enables the synthesis of complex outputs by composing simpler semantic components~\cite{du2023reduce}. Building on this compositional property, Du et al.~\cite{du2020compositional} leverage EBMs for image generation to satisfy logical concept constraints by composing individual attribute energies. This paradigm has been extended to representation learning and scene rearrangement~\cite{gkanatsios2023energy}. More recently, EDLM~\cite{xu2024energy} introduces sequence-level EBMs at each diffusion step to improve text generation, while EnergyMoGen~\cite{zhang2025energymogen} applies energy-based fusion in latent space to compose multiple semantic concepts for human motion generation. In contrast to these approaches, we propose EnergyAction, the first method to model bimanual manipulation policy as a composition of unimanual policies. By formulating pre-trained single-arm policies as energy functions and combining them through energy summation, we enable efficient compositional transfer to complex bimanual tasks with minimal bimanual data.

\section{Method}
\begin{figure*}[t] 
	\centering 
	\includegraphics[width=1.0\textwidth]{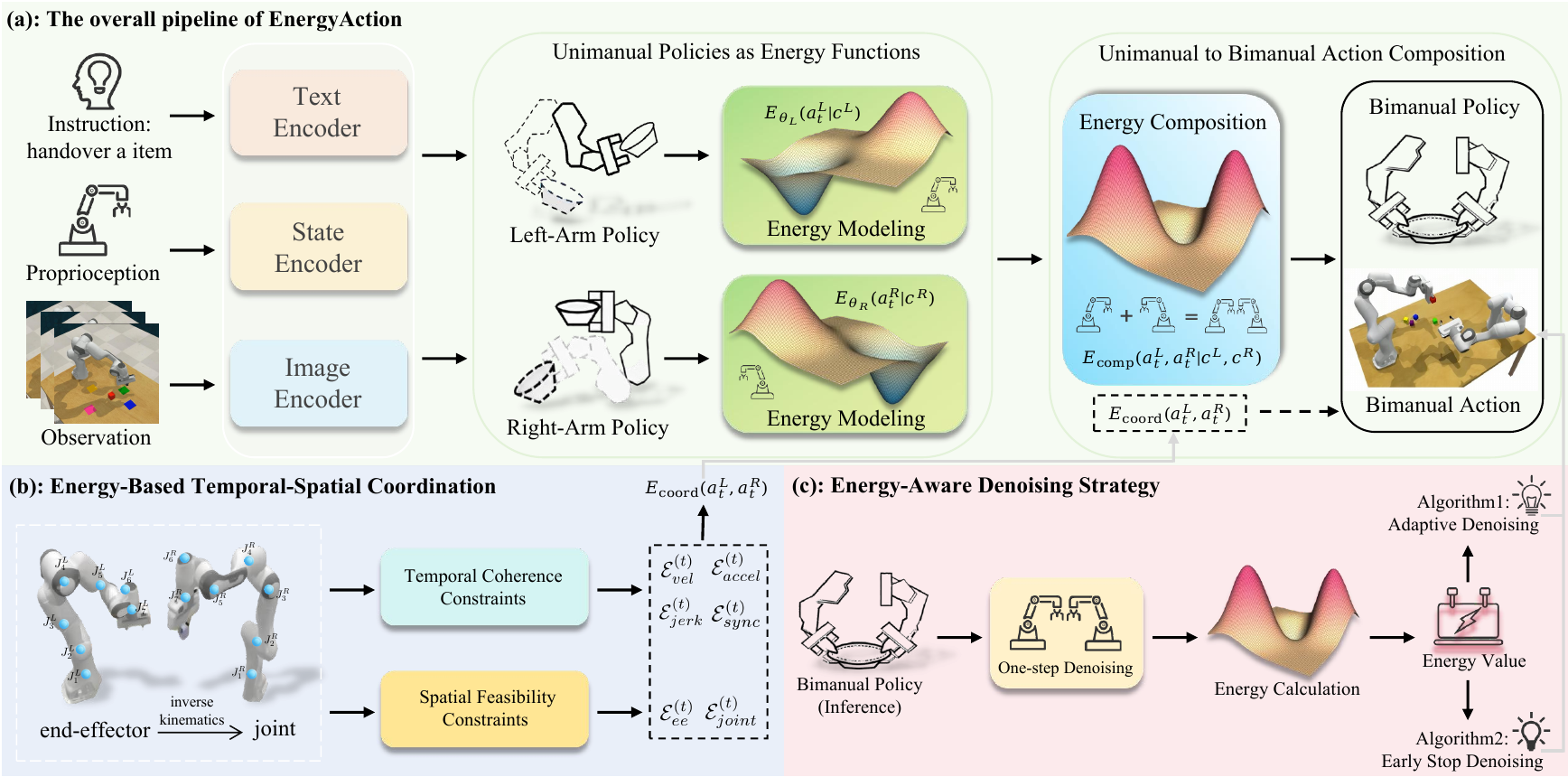} 
    \caption{\textbf{Overall of EnergyAction}. (a) We model unimanual policies as energy functions and compose them to form a bimanual policy. (b) To ensure coordinated bimanual actions, we introduce energy-based constraints from both temporal and spatial perspectives. (c) We propose two different energy-aware denoising strategies that adaptively adjust denoising steps based on energy values.}
	\label{fig:2} 
\end{figure*}

%%%%%%%%%%%%%%%%%%%%%%%%%%%%%%%%%%%%%%%%%%%%%%%%%%%%%%%%%%%%%%%%%%%%%%%%%%%%%%%%
%Method包括五部分
%第一部分是Preliminary，首先介绍了下双臂任务的基本设定和符号表示，然后简单介绍了下flow matching的基本原理以及如何应用于manipulation任务。
%第二部分是基于能量的组合动作生成。3.2.2从能量模型与flow matching的关系进行切入，然后利用能量模型把单臂的policy建模成能量模型的形式。有了单臂的能量模型之后，进入3.2.2，把单臂的能量模型组合成双臂（这部分涉及公式推导，主要是贝叶斯和一些取log的计算，放到了附录1）。
%第三部分从时间和空间的角度对组合生成的action进行约束（命名为协同能量），对应几个简单的公式。时间上分别是：速度，加速度，jerk以及双臂的方向和角度的同步约束。
%第四部分是组合能量与协同能量的组合优化，说明我们文章的训练优化方式。
%第五部分描述了我们提出的Energy-Aware Denoising Strategy，即推理时候可以把能量模型看做一个判别模型利用能量的大小判断生成动作的好坏，这部分对应两种不同的去噪方法（正文部分进行了文字的简单描述，具体伪代码放到了附录）。
%%%%%%%%%%%%%%%%%%%%%%%%%%%%%%%%%%%%%%%%%%%%%%%%%%%%%%%%%%%%%%%%%%%%%%%%%%%%%%%%
We present EnergyAction, a compositional framework for bimanual manipulation that composes pre-trained unimanual policies through EBMs. As shown in Figure~\ref{fig:2}, our key insight is formulating unimanual policies as energy functions that can be composed via energy summation. We introduce temporal-spatial coordination constraints to ensure physical feasibility. Additionally, we propose two different adaptive denoising strategies accelerate inference based on the magnitude of the energy values.
\subsection{Preliminary}
% \subsubsection{Energy-Based Models}
% EBMs model complex distributions through energy functions that assign lower energies to more probable configurations. For a state $x$, an EBM defines probability via the Boltzmann distribution:
% \begin{equation}
% p_\theta(x) = \frac{\exp(-E_\theta(x))}{Z(\theta)},
% \end{equation}
% where $E_\theta(x)$ is the energy function and $Z(\theta) = \int \exp(-E_\theta(x))dx$ is the partition function.

% The key compositional property of EBMs allows combining multiple energy functions through summation. Given functions $E_{\theta_1}(x)$ and $E_{\theta_2}(x)$, the joint distribution is:
% \begin{equation}
% p(x) \propto \exp(-(E_{\theta_1}(x) + E_{\theta_2}(x))).
% \end{equation}
% This enables modeling complex bimanual manipulation by combining unimanual energy functions.
\subsubsection{Task Definition}
We formulate bimanual manipulation as learning a policy $\pi$ that generates coordinated actions $a_t = (a_t^L, a_t^R)$ for both arms at each timestep $t$, conditioned on visual observations $o_t$, proprioceptive information $p_t$, and language instruction $l$. Each action specifies the 6D end-effector pose and gripper state. The conditioning information for each arm is composed as $c^i = (o_t, p_t^i, l)$ for $i \in \{L,R\}$, where $o_t$ is shared visual observations, $p_t^i$ is arm-specific proprioceptive state, and $l$ is the language instruction. 
\subsubsection{Flow Matching}
Flow matching~\cite{zhang2024robot,wang2025flowram} offers an efficient alternative to diffusion models for generative modeling. Unlike diffusion models that iteratively remove Gaussian noise, flow matching generates samples by solving an optimal transport problem between a source distribution $\mu_0$ and a target distribution $\mu_1$. The key idea is to construct a continuous trajectory $X = \{X_t : t \in [0, 1]\}$ that transforms samples from $\mu_0$ to $\mu_1$ by following straight paths. This trajectory is governed by an ordinary differential equation (ODE):
\begin{equation}
dX_t = v_t(X_t)dt, \quad t \in [0, 1], \quad X_0 \sim \mu_0,
\end{equation}
where $v_t$ is a time-dependent velocity field. The model learns to approximate the optimal velocity field by minimizing:
\begin{equation}
\mathcal{L}_\theta = \mathbb{E}_{t,X_1}\left[\|v_\theta(X_t, t) - v^*_t(X_t)\|^2\right],
\end{equation}
where $X_t = (1 - t)X_0 + tX_1$ denotes the linear interpolation at time $t$, and $v^*_t$ represents the optimal velocity field.

During inference, the model iteratively transforms a noise sample $X_0 \sim \mathcal{N}(0, I)$ into the predicted action $X_1$ over $N$ steps with step size $\Delta t = \frac{1}{N}$:
\begin{equation}
X_{t+\Delta t} = X_t + \Delta t \cdot v_\theta(X_t, t).
\end{equation}
This deterministic process significantly reduces computation during inference while maintaining high-quality generation, making it particularly suitable for real-time robotic applications.

\subsection{Compositional Action Generation with Energy}

Inspired by the theory of EBMs~\cite{du2020compositional,du2023reduce,xu2024energy}, we formulate bimanual action generation as an energy-based composition problem. We first interpret unimanual policies as energy functions and then compose them to obtain bimanual policy and generate bimanual actions.

\subsubsection{Unimanual Policies as Energy Functions} 

For a unimanual flow matching policy, we establish a connection between the unimanual policy and energy-based formulations to enable compositional generation. The inference process of flow matching follows a deterministic ODE: 
\begin{equation} 
a_{t+\Delta t}^i = a_t^i + \Delta t \cdot v_\theta(a_t^i, t, c^i), 
\end{equation} 
where $v_\theta(a_t^i, t, c^i)$ is the learned velocity field, and $t \in [0,1]$ denotes the continuous diffusion time. In EBMs, Langevin Dynamics-based 
sampling~\cite{lecun2006tutorial} follows:
\begin{equation} 
a_{k+1}^i = a_k^i - \eta \nabla_{a^i} E_\theta(a_k^i, c^i) + \sqrt{2\eta}\epsilon, 
\quad \epsilon \sim \mathcal{N}(0, I), 
\end{equation} 
where $k$ denotes discrete sampling steps, which is similar to the flow matching update. With the velocity field $v_\theta(a_t^i, t, c^i)$ approximated by $-\nabla_{a^i} E_\theta(a_t^i, c^i)$ in the deterministic limit (when the noise term vanishes and $\eta = \Delta t$), the update can be written as:
\begin{equation} 
a_{t+\Delta t}^i = a_t^i - \Delta t \cdot \nabla_{a^i} E_\theta(a_t^i, c^i). 
\end{equation}
By representing the energy function $E_\theta(a_t^i, c^i)$ through the velocity 
field $v_\theta$, flow matching policies can be interpreted as EBMs under the 
deterministic setting, establishing:
\begin{equation} 
v_\theta(a_t^i, t, c^i) = -\nabla_{a^i} E_\theta(a_t^i, c^i). 
\label{eq:velocity_energy} 
\end{equation}
Following the Boltzmann distribution formulation, we define the time-dependent 
energy function as: 
\begin{equation} 
E_\theta(a_t^i, c^i) := - \log p_t(a_t^i|c^i) + \text{const}, 
\end{equation} 
which satisfies $p_t(a_t^i|c^i) = \frac{\exp(-E_\theta(a_t^i, c^i))}{Z_t(\theta,c^i)}$, 
and $Z_t(\theta,c^i)$ is the partition function. This energy-based interpretation enables each unimanual policy to be viewed as an implicit energy function, facilitating compositional generation through the summation of energy functions from multiple policies.

\subsubsection{Unimanual to Bimanual Action Composition}

For bimanual tasks, we leverage two unimanual policies characterized 
by energy functions $E_{\theta_L}(a_t^L, c^L)$ and $E_{\theta_R}(a_t^R, c^R)$, 
where $c^L$ and $c^R$ represent conditioning information for each arm. To maintain the modular structure of pre-trained unimanual policies and enable efficient knowledge transfer, we construct the compositional energy by aggregating unimanual energies:
\begin{equation}
E_{comp}(a_t^L, a_t^R | c^L, c^R) = E_{\theta_L}(a_t^L, c^L) + E_{\theta_R}(a_t^R, c^R).
\label{eq:comp_energy}
\end{equation}
Drawing from the compositional generation principle in EBMs~\cite{du2020compositional}, 
the probability distribution of bimanual actions can be formulated as:
\begin{equation}
p(a_t|c^L, c^R) \propto p(a_t) \frac{p(a_t^L|c^L)}{p(a_t^L)} \frac{p(a_t^R|c^R)}{p(a_t^R)}.
\label{eq:comp_prob}
\end{equation}
This formulation enables us to interpret $p(a_t^i|c^i)$ and $p(a_t^i)$ as 
conditional and unconditional distributions, connecting to classifier-free 
guidance~\cite{ho2022classifier}. Taking the gradient of the log probability 
in Equation~\eqref{eq:comp_prob} and applying the velocity-energy relationship 
from Equation~\eqref{eq:velocity_energy}, we derive the compositional velocity 
field:
\begin{equation}
v_{comp}(a_t, t) = v_\theta(a_t, t) + \sum_{i \in \{L,R\}} w_i[v_\theta^i(a_t^i, t, c^i) - v_\theta^i(a_t^i, t)],
\end{equation}
where $v_\theta(a_t, t)$ is the unconditional bimanual velocity field, $v_\theta^i(a_t^i, t, c^i)$ represents the conditional velocity field for arm $i \in \{L,R\}$, and $w_i$ are guidance weights set to 1 in our implementation. This composition strategy enables coordinated bimanual action generation by combining unimanual policies, achieving effective compositional transfer of manipulation knowledge from unimanual to bimanual tasks.
\subsection{Energy-Based Temporal-Spatial Coordination}
While energy-based composition enables knowledge transfer from unimanual policies, generated bimanual actions may violate coordination constraints. We introduce energy-based coordination to enforce temporal and spatial constraints, ensuring generated bimanual actions are both temporal coherence and spatial feasibility. Additionally, we employ adaptive weights that automatically balance different objectives based on the current state.
\subsubsection{Temporal Coordination Constraints}

We enforce temporal coordination through smoothness constraints on the end-effector pose trajectories. Let $a_t^{i}$ denote the end-effector pose of arm $i \in \{L, R\}$ at timestep $t$. We encourage gentle motion by penalizing large velocities using the first-order finite difference:
\begin{equation}
\mathcal{E}_{vel}^{(t)} = \sum_{i \in \{L, R\}}\|a_t^{i} - a_{t-1}^{i}\|^2.
\end{equation}
To ensure smooth motion, we penalize acceleration using the second-order finite difference:
\begin{equation}
\mathcal{E}_{accel}^{(t)} = \sum_{i \in \{L, R\}}\|a_t^{i} - 2a_{t-1}^{i} + a_{t-2}^{i}\|^2.
\end{equation}
We further minimize jerk by penalizing the third-order finite difference of positions, which approximates the rate of change of acceleration:
\begin{equation}
\mathcal{E}_{jerk}^{(t)} = \sum_{i \in \{L, R\}}\|a_t^{i} - 3a_{t-1}^{i} + 3a_{t-2}^{i} - a_{t-3}^{i}\|^2.
\end{equation}
These three terms create a hierarchy of smoothness constraints spanning velocity, acceleration, and jerk.

For bimanual coordination, we require coherent motion between arms. Computing velocities as $v_t^{i} = a_t^{i} - a_{t-1}^{i}$, we penalize differences in both magnitude and direction:
\begin{equation}
\mathcal{E}_{sync}^{(t)} = \|\|v_t^{L}\|_2 - \|v_t^{R}\|_2\|^2 + \|\hat{v}_t^{L} - \hat{v}_t^{R}\|^2,
\end{equation}
where $\hat{v}_t^{i} = v_t^{i}/\|v_t^{i}\|_2$ denotes the normalized velocity direction.

\subsubsection{Spatial Feasibility Constraints}

We enforce spatial constraints to prevent inter-arm collisions. Let $\text{pos}(\cdot): \mathbb{R}^{7} \rightarrow \mathbb{R}^{3}$ denote the function that extracts the 3D position from the end-effector pose. We compute the Euclidean distance between the end-effectors:
\begin{equation}
d_{ee}(a_t^{L}, a_t^{R}) = \|\text{pos}(a_t^{L}) - \text{pos}(a_t^{R})\|_2,
\end{equation}
and define a smooth collision avoidance energy to prevent gradient discontinuities:
\begin{equation}
\mathcal{E}_{ee}^{(t)} = \max(0, d_{safe} - d_{ee}(a_t^{L}, a_t^{R}))^2,
\end{equation}
where $d_{safe}$ is the minimum safe distance and is set to 0.001 meters.

At the joint level, we use inverse kinematics~\cite{zhao2024inverse} to obtain joint configurations $j_t^{i} = \text{IK}(a_t^{i}, j_{t-1}^{i})$ and define an auxiliary constraint to encourage diverse configurations:
\begin{equation}
\mathcal{E}_{joint}^{(t)} = \max(0, d_{safe}^{joint} - \|j_t^{L} - j_t^{R}\|_2)^2,
\end{equation}
where $d_{safe}^{joint}$ represents the minimum safe distance in joint space, set to 0.001 meters. This smooth constraint formulation ensures continuous gradients throughout the optimization process.

\subsection{Unified Energy-Based Optimization}

We integrate compositional energy from unimanual policies and coordination constraints into a unified optimization framework. The coordination energy at timestep $t$ aggregates six constraint terms:
\begin{equation}
E_{coord}^{(t)}(a_t^{L}, a_t^{R}) = \sum_{i=1}^{6} w_i^{(t)} \cdot \mathcal{E}_{i}^{(t)},
\end{equation}
where $\mathcal{E}_{1}^{(t)}, \ldots, \mathcal{E}_{6}^{(t)}$ represent velocity, acceleration, jerk, synchronization, end-effector, and joint constraints respectively, and $w_i^{(t)}$ are adaptive weights balancing each constraint. To eliminate manual hyperparameter tuning, we use a lightweight MLP to predict weights from the current state:
\begin{equation}
\{w_1^{(t)}, \ldots, w_6^{(t)}\} = \text{softmax}(\text{MLP}([a_t^{L}; a_t^{R}; v_t^{L}; v_t^{R}])),
\end{equation}
where $[\cdot;\cdot]$ denotes concatenation. This enables adaptive constraint emphasis: collision avoidance when arms approach and smoothness when velocity changes occur.

The total energy function at timestep $t$ combines generation and coordination objectives:
\begin{equation}
E_{total}^{(t)} = E_{comp}^{(t)}(a_t^{L}, a_t^{R}) + E_{coord}^{(t)}(a_t^{L}, a_t^{R}),
\end{equation}
where $E_{comp}^{(t)}$ represents the compositional energy. The differentiability of the entire energy function enables end-to-end optimization to generate bimanual actions that are semantically meaningful (via compositional energies), smooth and coherent (via temporal constraints), and collision-free (via spatial constraints).

\subsection{Energy-Aware Denoising Strategy}
\renewcommand{\arraystretch}{0.97}
\begin{table*}[ht]
  \centering
  \setlength{\tabcolsep}{3pt}
  \caption{Results on RLBench2 with 20 and 100 demonstrations. Best results are in \textbf{bold}. We report the average success rate over 3 seeds.}
  \resizebox{\linewidth}{!}{
    \begin{tabular}{l|cc|cc|cc|cc|cc|cc|cc}
    \toprule[1pt]
    & \multicolumn{2}{c|}{Push Box} & \multicolumn{2}{c|}{Lift Ball} & \multicolumn{2}{c|}{Push Buttons} & \multicolumn{2}{c|}{Pick up Plate} & \multicolumn{2}{c|}{Put item into Drawer} & \multicolumn{2}{c|}{Put Bottle into Fridge} & \multicolumn{2}{c}{Handover item} \\
    \cmidrule(lr){2-3}\cmidrule(lr){4-5}\cmidrule(lr){6-7}\cmidrule(lr){8-9}\cmidrule(lr){10-11}\cmidrule(lr){12-13}\cmidrule(lr){14-15}
    Method & 20 & 100 & 20 & 100 & 20 & 100 & 20 & 100 & 20 & 100 & 20 & 100 & 20 & 100 \\
    \midrule
    ACT~\cite{zhao2023learning}  & 26.0\textcolor[gray]{0.5}{\small{$\pm4.6$}} & 48.7\textcolor[gray]{0.5}{\small{$\pm1.6$}} & 17.3\textcolor[gray]{0.5}{\small{$\pm5.8$}} & 40.7\textcolor[gray]{0.5}{\small{$\pm1.9$}} & 0.0\textcolor[gray]{0.5}{\small{$\pm0.0$}} & 4.0\textcolor[gray]{0.5}{\small{$\pm1.2$}} & 0.0\textcolor[gray]{0.5}{\small{$\pm0.0$}} & 0.0\textcolor[gray]{0.5}{\small{$\pm0.0$}} & 0.0\textcolor[gray]{0.5}{\small{$\pm0.0$}} & 13.0\textcolor[gray]{0.5}{\small{$\pm1.7$}} & 0.0\textcolor[gray]{0.5}{\small{$\pm0.0$}} & 0.0\textcolor[gray]{0.5}{\small{$\pm0.0$}} & 0.0\textcolor[gray]{0.5}{\small{$\pm0.0$}} & 0.0\textcolor[gray]{0.5}{\small{$\pm0.0$}} \\
    RVT-LF~\cite{goyal2023rvt,grotz2024peract2} & 44.0\textcolor[gray]{0.5}{\small{$\pm1.0$}} & 76.3\textcolor[gray]{0.5}{\small{$\pm1.5$}} & 12.3\textcolor[gray]{0.5}{\small{$\pm6.7$}} & 28.7\textcolor[gray]{0.5}{\small{$\pm1.0$}} & 12.5\textcolor[gray]{0.5}{\small{$\pm3.5$}} & 39.0\textcolor[gray]{0.5}{\small{$\pm1.7$}} & 1.0\textcolor[gray]{0.5}{\small{$\pm0.0$}} & 3.0\textcolor[gray]{0.5}{\small{$\pm1.8$}} & 0.0\textcolor[gray]{0.5}{\small{$\pm0.0$}} & 10.0\textcolor[gray]{0.5}{\small{$\pm1.0$}} & 0.0\textcolor[gray]{0.5}{\small{$\pm0.0$}} & 0.0\textcolor[gray]{0.5}{\small{$\pm0.0$}} & 0.7\textcolor[gray]{0.5}{\small{$\pm0.5$}} & 0.0\textcolor[gray]{0.5}{\small{$\pm0.0$}} \\
    PerAct-LF~\cite{shridhar2023perceiver,grotz2024peract2} & 63.7\textcolor[gray]{0.5}{\small{$\pm3.1$}} & 66.3\textcolor[gray]{0.5}{\small{$\pm3.1$}} & 42.3\textcolor[gray]{0.5}{\small{$\pm8.1$}} & 68.3\textcolor[gray]{0.5}{\small{$\pm6.1$}} & 8.0\textcolor[gray]{0.5}{\small{$\pm2.8$}} & 30.0\textcolor[gray]{0.5}{\small{$\pm3.5$}} & 3.0\textcolor[gray]{0.5}{\small{$\pm0.0$}} & 4.0\textcolor[gray]{0.5}{\small{$\pm0.9$}} & 27.0\textcolor[gray]{0.5}{\small{$\pm2.5$}} & 27.0\textcolor[gray]{0.5}{\small{$\pm0.7$}} & 2.0\textcolor[gray]{0.5}{\small{$\pm0.3$}} & 7.0\textcolor[gray]{0.5}{\small{$\pm0.0$}} & 0.3\textcolor[gray]{0.5}{\small{$\pm0.3$}} & 3.0\textcolor[gray]{0.5}{\small{$\pm0.0$}} \\
    PerAct2~\cite{grotz2024peract2} & 59.7\textcolor[gray]{0.5}{\small{$\pm7.6$}} & 77.0\textcolor[gray]{0.5}{\small{$\pm3.0$}} & 37.7\textcolor[gray]{0.5}{\small{$\pm4.7$}} & 50.0\textcolor[gray]{0.5}{\small{$\pm3.2$}} & 50.7\textcolor[gray]{0.5}{\small{$\pm4.2$}} & 47.0\textcolor[gray]{0.5}{\small{$\pm2.2$}} & 2.0\textcolor[gray]{0.5}{\small{$\pm1.8$}} & 4.0\textcolor[gray]{0.5}{\small{$\pm0.0$}} & 2.3\textcolor[gray]{0.5}{\small{$\pm1.5$}} & 10.0\textcolor[gray]{0.5}{\small{$\pm0.9$}} & 7.0\textcolor[gray]{0.5}{\small{$\pm0.0$}} & 16.0\textcolor[gray]{0.5}{\small{$\pm2.4$}} & 17.3\textcolor[gray]{0.5}{\small{$\pm0.6$}} & 11.0\textcolor[gray]{0.5}{\small{$\pm1.5$}} \\
    DP3~\cite{ze20243d}  & 25.3\textcolor[gray]{0.5}{\small{$\pm4.0$}} & 56.0\textcolor[gray]{0.5}{\small{$\pm3.6$}} & 34.3\textcolor[gray]{0.5}{\small{$\pm7.6$}} & 64.0\textcolor[gray]{0.5}{\small{$\pm2.7$}} & 18.0\textcolor[gray]{0.5}{\small{$\pm2.3$}} & 35.0\textcolor[gray]{0.5}{\small{$\pm2.5$}} & 8.5\textcolor[gray]{0.5}{\small{$\pm1.8$}} & 22.0\textcolor[gray]{0.5}{\small{$\pm2.1$}} & 12.0\textcolor[gray]{0.5}{\small{$\pm2.0$}} & 28.5\textcolor[gray]{0.5}{\small{$\pm2.3$}} & 5.5\textcolor[gray]{0.5}{\small{$\pm1.5$}} & 18.5\textcolor[gray]{0.5}{\small{$\pm2.0$}} & 3.2\textcolor[gray]{0.5}{\small{$\pm1.2$}} & 12.0\textcolor[gray]{0.5}{\small{$\pm1.8$}} \\
    KStarDiffuser~\cite{lv2025spatial} & 79.3\textcolor[gray]{0.5}{\small{$\pm3.5$}} & 83.0\textcolor[gray]{0.5}{\small{$\pm1.7$}} & 87.0\textcolor[gray]{0.5}{\small{$\pm2.7$}} & 98.7\textcolor[gray]{0.5}{\small{$\pm1.5$}} & 28.0\textcolor[gray]{0.5}{\small{$\pm1.2$}} & 38.0\textcolor[gray]{0.5}{\small{$\pm1.6$}} & 5.5\textcolor[gray]{0.5}{\small{$\pm0.9$}} & 10.0\textcolor[gray]{0.5}{\small{$\pm1.4$}} & 9.0\textcolor[gray]{0.5}{\small{$\pm0.9$}} & 15.0\textcolor[gray]{0.5}{\small{$\pm1.3$}} & 9.5\textcolor[gray]{0.5}{\small{$\pm0.8$}} & 16.0\textcolor[gray]{0.5}{\small{$\pm2.0$}} & 5.8\textcolor[gray]{0.5}{\small{$\pm0.6$}} & 14.5\textcolor[gray]{0.5}{\small{$\pm1.4$}} \\
    AnyBimanual~\cite{lu2025anybimanual} & 31.0\textcolor[gray]{0.5}{\small{$\pm2.1$}} & 46.0\textcolor[gray]{0.5}{\small{$\pm2.6$}} & 22.0\textcolor[gray]{0.5}{\small{$\pm2.1$}} & 36.0\textcolor[gray]{0.5}{\small{$\pm2.0$}} & 39.0\textcolor[gray]{0.5}{\small{$\pm2.4$}} & 73.0\textcolor[gray]{0.5}{\small{$\pm2.3$}} & 6.0\textcolor[gray]{0.5}{\small{$\pm2.1$}} & 8.0\textcolor[gray]{0.5}{\small{$\pm2.2$}} & 16.0\textcolor[gray]{0.5}{\small{$\pm2.4$}} & 32.0\textcolor[gray]{0.5}{\small{$\pm2.5$}} & 13.0\textcolor[gray]{0.5}{\small{$\pm2.0$}} & 26.0\textcolor[gray]{0.5}{\small{$\pm2.7$}} & 7.0\textcolor[gray]{0.5}{\small{$\pm2.5$}} & 15.0\textcolor[gray]{0.5}{\small{$\pm2.9$}} \\
    $\pi_0$-keypose~\cite{black2410pi0} & 55.0\textcolor[gray]{0.5}{\small{$\pm1.9$}} & 93.0\textcolor[gray]{0.5}{\small{$\pm1.6$}} & 68.0\textcolor[gray]{0.5}{\small{$\pm2.4$}} & 97.0\textcolor[gray]{0.5}{\small{$\pm1.9$}} & 36.0\textcolor[gray]{0.5}{\small{$\pm1.9$}} & 58.0\textcolor[gray]{0.5}{\small{$\pm1.8$}} & 39.0\textcolor[gray]{0.5}{\small{$\pm2.0$}} & 47.0\textcolor[gray]{0.5}{\small{$\pm1.8$}} & 38.0\textcolor[gray]{0.5}{\small{$\pm2.0$}} & 44.0\textcolor[gray]{0.5}{\small{$\pm0.8$}} & 20.0\textcolor[gray]{0.5}{\small{$\pm2.3$}} & 38.0\textcolor[gray]{0.5}{\small{$\pm1.3$}} & 1.0\textcolor[gray]{0.5}{\small{$\pm1.6$}} & 33.0\textcolor[gray]{0.5}{\small{$\pm1.6$}} \\
    3DFA~\cite{gkanatsios20253d} & 70.0\textcolor[gray]{0.5}{\small{$\pm2.0$}} & 84.0\textcolor[gray]{0.5}{\small{$\pm1.2$}} & 84.0\textcolor[gray]{0.5}{\small{$\pm2.2$}} & \textbf{99.0}\textcolor[gray]{0.5}{\small{$\pm0.6$}} & 73.0\textcolor[gray]{0.5}{\small{$\pm1.8$}} & 92.0\textcolor[gray]{0.5}{\small{$\pm1.5$}} & 53.0\textcolor[gray]{0.5}{\small{$\pm2.1$}} & 59.0\textcolor[gray]{0.5}{\small{$\pm2.3$}} & 53.0\textcolor[gray]{0.5}{\small{$\pm1.9$}} & 88.0\textcolor[gray]{0.5}{\small{$\pm1.8$}} & 49.0\textcolor[gray]{0.5}{\small{$\pm2.3$}} & 89.0\textcolor[gray]{0.5}{\small{$\pm1.4$}} & 43.0\textcolor[gray]{0.5}{\small{$\pm1.6$}} & \textbf{91.0}\textcolor[gray]{0.5}{\small{$\pm0.9$}} \\
    \rowcolor{blue!5!cyan!10}EnergyAction (Ours) & \textbf{81.0}\textcolor[gray]{0.5}{\small{$\pm0.0$}} & \textbf{94.3}\textcolor[gray]{0.5}{\small{$\pm1.5$}} & \textbf{98.0}\textcolor[gray]{0.5}{\small{$\pm2.0$}} & 98.3\textcolor[gray]{0.5}{\small{$\pm1.2$}} & \textbf{82.5}\textcolor[gray]{0.5}{\small{$\pm3.1$}} & \textbf{93.0}\textcolor[gray]{0.5}{\small{$\pm1.0$}} & \textbf{72.0}\textcolor[gray]{0.5}{\small{$\pm4.8$}} & \textbf{73.7}\textcolor[gray]{0.5}{\small{$\pm3.2$}} & \textbf{92.0}\textcolor[gray]{0.5}{\small{$\pm0.8$}} & \textbf{93.3}\textcolor[gray]{0.5}{\small{$\pm3.1$}} & \textbf{50.0}\textcolor[gray]{0.5}{\small{$\pm8.5$}} & \textbf{89.3}\textcolor[gray]{0.5}{\small{$\pm4.7$}} & \textbf{68.0}\textcolor[gray]{0.5}{\small{$\pm7.0$}} & 87.3\textcolor[gray]{0.5}{\small{$\pm4.5$}} \\
    \midrule[0.75pt]
    & \multicolumn{2}{c|}{Pick up Laptop} & \multicolumn{2}{c|}{Straighten Rope} & \multicolumn{2}{c|}{Sweep Dust} & \multicolumn{2}{c|}{Lift Tray} & \multicolumn{2}{c|}{Handover item (easy)} & \multicolumn{2}{c|}{Take Tray out of Oven} & \multicolumn{2}{c}{\cellcolor[rgb]{.902,.996,1.0}Avg. Success}\\
    \cmidrule(lr){2-3}\cmidrule(lr){4-5}\cmidrule(lr){6-7}\cmidrule(lr){8-9}\cmidrule(lr){10-11}\cmidrule(lr){12-13}\cmidrule(lr){14-15}
    & 20 & 100 & 20 & 100 & 20 & 100 & 20 & 100 & 20 & 100 & 20 & 100 & 20 & 100 \\
    \midrule
   ACT~\cite{zhao2023learning}  & 0.0\textcolor[gray]{0.5}{\small{$\pm0.0$}} & 0.0\textcolor[gray]{0.5}{\small{$\pm0.0$}} & 0.0\textcolor[gray]{0.5}{\small{$\pm0.0$}} & 16.0\textcolor[gray]{0.5}{\small{$\pm1.5$}} & 0.0\textcolor[gray]{0.5}{\small{$\pm0.0$}} & 0.0\textcolor[gray]{0.5}{\small{$\pm0.0$}} & 0.0\textcolor[gray]{0.5}{\small{$\pm0.0$}} & 6.0\textcolor[gray]{0.5}{\small{$\pm1.4$}} & 0.0\textcolor[gray]{0.5}{\small{$\pm0.0$}} & 0.0\textcolor[gray]{0.5}{\small{$\pm0.0$}} & 0.0\textcolor[gray]{0.5}{\small{$\pm0.0$}} & 2.0\textcolor[gray]{0.5}{\small{$\pm1.7$}} & 3.3\textcolor[gray]{0.5}{\small{$\pm0.8$}} & 10.0\textcolor[gray]{0.5}{\small{$\pm0.8$}} \\
    RVT-LF~\cite{goyal2023rvt,grotz2024peract2} & 1.0\textcolor[gray]{0.5}{\small{$\pm0.8$}} & 3.0\textcolor[gray]{0.5}{\small{$\pm1.2$}} & 0.5\textcolor[gray]{0.5}{\small{$\pm0.4$}} & 3.0\textcolor[gray]{0.5}{\small{$\pm2.0$}} & 0.0\textcolor[gray]{0.5}{\small{$\pm0.0$}} & 2.3\textcolor[gray]{0.5}{\small{$\pm1.7$}} & 4.0\textcolor[gray]{0.5}{\small{$\pm1.2$}} & 6.0\textcolor[gray]{0.5}{\small{$\pm0.7$}} & 3.0\textcolor[gray]{0.5}{\small{$\pm0.4$}} & 3.0\textcolor[gray]{0.5}{\small{$\pm0.2$}} & 2.0\textcolor[gray]{0.5}{\small{$\pm0.0$}} & 3.0\textcolor[gray]{0.5}{\small{$\pm1.2$}} & 6.2\textcolor[gray]{0.5}{\small{$\pm1.1$}} & 13.6\textcolor[gray]{0.5}{\small{$\pm1.1$}} \\
    PerAct-LF~\cite{grotz2024peract2,shridhar2023perceiver} & 7.0\textcolor[gray]{0.5}{\small{$\pm2.7$}} & 14.3\textcolor[gray]{0.5}{\small{$\pm3.8$}} & 4.3\textcolor[gray]{0.5}{\small{$\pm3.1$}} & 21.0\textcolor[gray]{0.5}{\small{$\pm0.8$}} & 22.0\textcolor[gray]{0.5}{\small{$\pm0.6$}} & 47.0\textcolor[gray]{0.5}{\small{$\pm2.7$}} & 6.0\textcolor[gray]{0.5}{\small{$\pm0.0$}} & 14.0\textcolor[gray]{0.5}{\small{$\pm1.1$}} & 3.3\textcolor[gray]{0.5}{\small{$\pm0.6$}} & 9.0\textcolor[gray]{0.5}{\small{$\pm2.1$}} & 1.0\textcolor[gray]{0.5}{\small{$\pm0.0$}} & 8.0\textcolor[gray]{0.5}{\small{$\pm1.3$}} & 14.6\textcolor[gray]{0.5}{\small{$\pm1.9$}} & 24.5\textcolor[gray]{0.5}{\small{$\pm2.0$}} \\
    PerAct2~\cite{grotz2024peract2} & 3.0\textcolor[gray]{0.5}{\small{$\pm0.6$}} & 12.0\textcolor[gray]{0.5}{\small{$\pm1.0$}} & 6.0\textcolor[gray]{0.5}{\small{$\pm0.7$}} & 24.0\textcolor[gray]{0.5}{\small{$\pm2.1$}} & 25.0\textcolor[gray]{0.5}{\small{$\pm1.5$}} & 52.0\textcolor[gray]{0.5}{\small{$\pm3.3$}} & 4.0\textcolor[gray]{0.5}{\small{$\pm2.1$}} & 5.0\textcolor[gray]{0.5}{\small{$\pm0.2$}} & 22.0\textcolor[gray]{0.5}{\small{$\pm1.0$}} & 41.0\textcolor[gray]{0.5}{\small{$\pm1.7$}} & 0.0\textcolor[gray]{0.5}{\small{$\pm0.0$}} & 9.0\textcolor[gray]{0.5}{\small{$\pm1.1$}} & 18.2\textcolor[gray]{0.5}{\small{$\pm2.0$}} & 27.5\textcolor[gray]{0.5}{\small{$\pm1.7$}} \\
    DP3~\cite{ze20243d}  & 2.7\textcolor[gray]{0.5}{\small{$\pm2.9$}} & 6.3\textcolor[gray]{0.5}{\small{$\pm3.1$}} & 2.0\textcolor[gray]{0.5}{\small{$\pm1.1$}} & 8.5\textcolor[gray]{0.5}{\small{$\pm1.6$}} & 0.0\textcolor[gray]{0.5}{\small{$\pm0.0$}} & 1.7\textcolor[gray]{0.5}{\small{$\pm2.1$}} & 8.0\textcolor[gray]{0.5}{\small{$\pm1.7$}} & 25.0\textcolor[gray]{0.5}{\small{$\pm2.2$}} & 0.0\textcolor[gray]{0.5}{\small{$\pm0.0$}} & 0.0\textcolor[gray]{0.5}{\small{$\pm0.0$}} & 6.5\textcolor[gray]{0.5}{\small{$\pm1.4$}} & 20.0\textcolor[gray]{0.5}{\small{$\pm1.9$}} & 9.7\textcolor[gray]{0.5}{\small{$\pm2.1$}} & 22.9\textcolor[gray]{0.5}{\small{$\pm2.1$}} \\
    KStarDiffuser~\cite{lv2025spatial} & 17.0\textcolor[gray]{0.5}{\small{$\pm2.0$}} & 43.7\textcolor[gray]{0.5}{\small{$\pm4.5$}} & 0.5\textcolor[gray]{0.5}{\small{$\pm0.4$}} & 3.5\textcolor[gray]{0.5}{\small{$\pm1.2$}} & 83.0\textcolor[gray]{0.5}{\small{$\pm4.4$}} & 89.0\textcolor[gray]{0.5}{\small{$\pm5.2$}} & 12.0\textcolor[gray]{0.5}{\small{$\pm1.9$}} & 28.0\textcolor[gray]{0.5}{\small{$\pm2.7$}} & 23.7\textcolor[gray]{0.5}{\small{$\pm0.6$}} & 27.0\textcolor[gray]{0.5}{\small{$\pm1.7$}} & 6.7\textcolor[gray]{0.5}{\small{$\pm1.9$}} & 22.0\textcolor[gray]{0.5}{\small{$\pm1.6$}} & 28.2\textcolor[gray]{0.5}{\small{$\pm1.7$}} & 37.6\textcolor[gray]{0.5}{\small{$\pm2.1$}} \\
    AnyBimanual~\cite{lu2025anybimanual} & 4.0\textcolor[gray]{0.5}{\small{$\pm2.3$}} & 7.0\textcolor[gray]{0.5}{\small{$\pm2.1$}} & 17.0\textcolor[gray]{0.5}{\small{$\pm2.9$}} & 24.0\textcolor[gray]{0.5}{\small{$\pm2.4$}} & 43.0\textcolor[gray]{0.5}{\small{$\pm2.2$}} & 67.0\textcolor[gray]{0.5}{\small{$\pm2.0$}} & 9.0\textcolor[gray]{0.5}{\small{$\pm2.3$}} & 14.0\textcolor[gray]{0.5}{\small{$\pm2.2$}} & 31.0\textcolor[gray]{0.5}{\small{$\pm2.3$}} & 44.0\textcolor[gray]{0.5}{\small{$\pm2.5$}} & 9.0\textcolor[gray]{0.5}{\small{$\pm2.2$}} & 24.0\textcolor[gray]{0.5}{\small{$\pm2.0$}} & 19.0\textcolor[gray]{0.5}{\small{$\pm2.3$}} & 32.0\textcolor[gray]{0.5}{\small{$\pm2.3$}} \\
    $\pi_0$-keypose~\cite{black2410pi0} & 22.0\textcolor[gray]{0.5}{\small{$\pm1.3$}} & 27.0\textcolor[gray]{0.5}{\small{$\pm1.2$}} & 6.0\textcolor[gray]{0.5}{\small{$\pm1.0$}} & 7.0\textcolor[gray]{0.5}{\small{$\pm1.9$}} & 1.0\textcolor[gray]{0.5}{\small{$\pm0.8$}} & 15.0\textcolor[gray]{0.5}{\small{$\pm1.1$}} & 27.0\textcolor[gray]{0.5}{\small{$\pm1.7$}} & 72.0\textcolor[gray]{0.5}{\small{$\pm2.1$}} & 37.0\textcolor[gray]{0.5}{\small{$\pm1.8$}} & 59.0\textcolor[gray]{0.5}{\small{$\pm1.8$}} & 10.0\textcolor[gray]{0.5}{\small{$\pm1.2$}} & 68.0\textcolor[gray]{0.5}{\small{$\pm1.7$}} & 27.7\textcolor[gray]{0.5}{\small{$\pm1.7$}} & 50.6\textcolor[gray]{0.5}{\small{$\pm1.6$}} \\
    3DFA~\cite{gkanatsios20253d} & 29.0\textcolor[gray]{0.5}{\small{$\pm1.5$}} & \textbf{70.0}\textcolor[gray]{0.5}{\small{$\pm2.1$}} & 9.0\textcolor[gray]{0.5}{\small{$\pm1.1$}} & 18.0\textcolor[gray]{0.5}{\small{$\pm1.6$}} & 19.0\textcolor[gray]{0.5}{\small{$\pm1.4$}} & 96.0\textcolor[gray]{0.5}{\small{$\pm0.8$}} & 39.0\textcolor[gray]{0.5}{\small{$\pm1.8$}} & \textbf{97.0}\textcolor[gray]{0.5}{\small{$\pm1.1$}} & 49.0\textcolor[gray]{0.5}{\small{$\pm1.7$}} & 89.0\textcolor[gray]{0.5}{\small{$\pm1.3$}} & 13.0\textcolor[gray]{0.5}{\small{$\pm1.2$}} & 92.0\textcolor[gray]{0.5}{\small{$\pm1.7$}} & 44.8\textcolor[gray]{0.5}{\small{$\pm1.7$}} & 81.8\textcolor[gray]{0.5}{\small{$\pm1.4$}} \\
    \rowcolor[rgb]{.902,.996,1.0}EnergyAction (Ours) & \textbf{43.5}\textcolor[gray]{0.5}{\small{$\pm8.7$}} & 63.3\textcolor[gray]{0.5}{\small{$\pm5.1$}} & \textbf{46.8}\textcolor[gray]{0.5}{\small{$\pm3.2$}} & \textbf{50.3}\textcolor[gray]{0.5}{\small{$\pm2.9$}} & \textbf{97.5}\textcolor[gray]{0.5}{\small{$\pm0.6$}} & \textbf{99.0}\textcolor[gray]{0.5}{\small{$\pm1.0$}} & \textbf{94.0}\textcolor[gray]{0.5}{\small{$\pm1.8$}} & 94.3\textcolor[gray]{0.5}{\small{$\pm2.1$}} & \textbf{89.8}\textcolor[gray]{0.5}{\small{$\pm2.8$}} & \textbf{95.0}\textcolor[gray]{0.5}{\small{$\pm2.0$}} & \textbf{90.0}\textcolor[gray]{0.5}{\small{$\pm2.2$}} & \textbf{92.7}\textcolor[gray]{0.5}{\small{$\pm0.6$}} & \textbf{77.3}\textcolor[gray]{0.5}{\small{$\pm3.5$}} & \textbf{86.4}\textcolor[gray]{0.5}{\small{$\pm2.5$}} \\
    \bottomrule[1pt]
    \end{tabular}
    }%
  \label{tab:simulation_performance}%
\end{table*}
% Our energy-based framework offers dual functionality: during training, it serves as a generative model for action distribution, while at inference, it acts as a discriminative critic to evaluate action quality. 
% During inference, starting from noise $a_0 \sim \mathcal{N}(0, I)$, we iteratively update the bimanual action $a_t = (a_t^L, a_t^R)$ over $N$ steps with step size $\Delta t = \frac{1}{N}$:
% \begin{equation}
% a_{t+\Delta t} = a_t + \Delta t \cdot v_{comp}(a_t, t),
% \label{eq:inference_update}
% \end{equation}
% to obtain the final prediction $a_1$. However, since different tasks or operations have varying levels of difficulty, a natural question arises: \emph{rather than applying uniform denoising steps across all operations, can we adapt the number of denoising steps based on the action difficulty?} During the inference phase, we introduce two different energy-aware adaptive denoising strategies according to energy assessment. Concretely, simple actions (low energy) require fewer denoising steps, while complex ones (high energy) require more. By using $E_{total}^{(t)}$ as a quality indicator, our strategies improve inference efficiency while ensuring action generation quality. We implement this idea through two complementary algorithms: 

During inference, starting from noise $a_0 \sim \mathcal{N}(0, I)$, we iteratively update the bimanual action $a_t = (a_t^L, a_t^R)$ over $N$ steps with step size $\Delta t = \frac{1}{N}$:
\begin{equation}
a_{t+\Delta t} = a_t + \Delta t \cdot v_{total}(a_t, t),
\label{eq:inference_update}
\end{equation}
where the total velocity field $v_{total}$ integrates compositional generation and coordination constraints. 
% However, since different tasks or operations have varying levels of difficulty, a natural question arises: \emph{rather than applying uniform denoising steps across all operations, can we adapt the number of denoising steps based on the action difficulty?}
To improve inference efficiency, we introduce two energy-aware adaptive denoising strategies that adjust denoising steps based on $\smash{E_{total}^{(t)}}$. Simple actions (low energy) require fewer steps, while complex ones (high energy) require more, implemented through two complementary algorithms:

\textbf{Algorithm 1 (Adaptive Denoising):} 
This strategy determines the number of denoising steps based on energy value. If $\smash{E_{total}^{(t)}} < \smash{\tau_{low}}$, we perform one denoising step (as we find that simple tasks succeed with single-step denoising); if $\smash{E_{total}^{(t)™}} > \smash{\tau_{high}}$, we use the maximum steps (e.g., 5 for flow matching); otherwise, we linearly interpolate between these thresholds. The thresholds $\smash{\tau_{low} = 4}$ and $\smash{\tau_{high} = 10}$ are set based on the energy distribution over training data.

\textbf{Algorithm 2 (Early Stop Denoising):} This strategy performs denoising with real-time energy monitoring. The denoising process terminates early once the energy falls below $\tau_{low}$, providing more flexible adaptation to varying action complexities. 
% We provide more detailed explanations and pseudocode in the Appendix~\ref{sec:Pseudocode}.

% For notation simplicity in the following algorithms, we denote ${a}_t = ({a}_t^{L}, {a}_t^{R})$ as the bimanual action. All subsequent operations, including velocity prediction and energy gradient computation, are performed jointly on both arms. We provide pseudocode in the Appendix~\ref{sec:Pseudocode}.

% In our experiments, we empirically set the energy thresholds as $\tau_{low}=4.0$ and $\tau_{high}=10.0$, with a maximum number of steps $N_{max}=10$. These parameters effectively balance computational efficiency and action quality across different manipulation scenarios.

\section{Experiments}
%%%%%%%%%%%%%%%%%%%%%%%%%%%%%%%%%%%%%%%%%%%%%%%%%%%%%%%%%%%%%%%%%%%%%%%%%%%%%%%%
%Experiments包括四部分
%第一部分是数据集和实验设定。仿真实验是rlbench1训练单臂任务，然后迁移到rlbench2（与anybimanual采用相同的设定），真实场景包括星海图的机械臂和aloha（aloha预计下周一到，放到附录）。
%第二部分是与现有的sota方法进行对比，包括basline的介绍以及实验结果的对比，分析我们的方法为什么比其他方法好（分而治之的思想，与introducion开头部分笛卡尔的名言呼应）。
%第三部分是消融实验，这部分尽可能多的展示不同demo数量设定下的实验结果，证明我们班的方法是data-effective的。尤其是比较少的时候，我们的效果尤其由于其他方法（与我们的动机一致）。
%同时做了：1.去噪策略的分析(说明我们的方法所需要的去噪步数少，推理更快)。
%2.单臂任务数量影响的实验（说明我们的方法哪怕不使用单臂任务预训练，也能优于其他方法，证明了我们基于能量模型组合建模的优势）。
%3.不同的扩散策略（包括ddpm，ddim），证明了我们的方法适用于任何的基于概率分布预测的模型，说明了我们方法的robustness and generalizability。
%第四部分是真实场景的实验，图片中是定性结果，表4是定量结果。
%%%%%%%%%%%%%%%%%%%%%%%%%%%%%%%%%%%%%%%%%%%%%%%%%%%%%%%%%%%%%%%%%%%%%%%%%%%%%%%%
\subsection{Datasets and Experimental Setup}
Following AnyBimanual~\cite{lu2025anybimanual} experimental setting, we train unimanual policies on RLBench~\cite{shridhar2023perceiver} and compositionally transfer them to bimanual tasks in RLBench2~\cite{grotz2024peract2}. RLBench provides 18 single-arm manipulation tasks, while RLBench2 extends this with 13 language-conditioned bimanual tasks requiring diverse coordination patterns. Visual observations are captured from six RGB-D cameras at 256$\times$256 resolution. To evaluate data efficiency, we conduct experiments with varying numbers of demonstrations (1, 5, 10, 20, and 100) during training, with each policy evaluated over 25 episodes per task. 
For real-world validation, we deploy different policies on the Galaxea R1 lite robot. 
% Please refer to Appendix~\ref{sec:sim_env} and~\ref{sec:real_world} for more details about simulated and real-world experimental settings.
\subsection{Comparison Results with SOTA Methods}
\textbf{Baselines.} We systematically evaluate EnergyAction against state-of-the-art methods including ACT~\cite{zhao2023learning}, RVT-LF~\cite{goyal2023rvt,grotz2024peract2}, PerAct-LF~\cite{shridhar2023perceiver,grotz2024peract2}, and PerAct$^2$~\cite{grotz2024peract2} (results from~\cite{grotz2024peract2}); AnyBimanual~\cite{lu2025anybimanual}; 3D Diffusion Policy (DP3)~\cite{ze20243d} and KStarDiffuser~\cite{lv2025spatial} (results from~\cite{lv2025spatial});  $\pi_0$~\cite{black2410pi0}, a variant that predicts 3D end-effector keyposes rather than joint angles, providing a standardized output format across all policies for improved performance; and 3D FlowMatch Actor (3DFA)~\cite{gkanatsios20253d} (100 demo results are reproduced by evaluating the publicly available checkpoint from 3DFA~\cite{gkanatsios20253d}). For methods that do not report results on certain tasks, we re-implement them and report the average success rate.\\
\textbf{Results.} Table~\ref{tab:simulation_performance} presents a quantitative comparison of EnergyAction against state-of-the-art methods across different data regimes. Our approach achieves the highest overall performance, with average success rates of 77.3\% and 86.4\% under the 20 and 100 demonstration settings respectively. Specifically, with 100 demonstrations, EnergyAction reaches an average success rate of 86.4\%, outperforming the strongest baseline 3DFA by 4.6\%. Notably, under the limited-data setting with only 20 demonstrations, our method maintains a substantial performance advantage of 32.5\% over the second-best approach, demonstrating superior data efficiency. 

The superior performance of EnergyAction stems from its ability to compositionally transfer knowledge from unimanual policies to bimanual manipulation policy, which effectively alleviates the optimization challenge posed by the high-dimensional bimanual action space. By decomposing the complex bimanual coordination problem into manageable unimanual components and subsequently composing them through our energy-based framework, our method follows a divide-and-conquer paradigm~\cite{jiang2025rethinking} that tackles the inherent complexity of bimanual manipulation. This decomposition strategy enables more efficient learning and better generalization, particularly in tasks requiring intricate coordination patterns. 

To further validate the effectiveness of our proposed energy-based framework, we conduct additional experiments with 1, 5, and 10 training demonstrations. The results consistently demonstrate that EnergyAction achieves superior performance across all low-data regimes, confirming the robustness and generalizability of our EnergyAction. 
% More detailed analysis are provided in Appendix~\ref{sec:ex_and_de}.

\subsection{Ablation Study}
% \renewcommand{\arraystretch}{1.0}
% \begin{table}[t]
%   \centering
%   \setlength{\tabcolsep}{8pt}
%   \caption{Ablation study of EnergyAction components.}
%   \resizebox{0.95\linewidth}{!}{
%     \begin{tabular}{ccc|cc}
%     \toprule[1pt]
%           \multirow{2}{*}{E-Compose} & \multirow{2}{*}{E-Temporal} & \multirow{2}{*}{E-Spatial}& \multicolumn{2}{c}{Avg. Success} \\
%     % \cmidrule(lr){4-5}
%           &  &  & 20-demo & 100-demo \\
%     \midrule
%     \rowcolor{blue!5!cyan!10}$\checkmark$ & $\checkmark$ & $\checkmark$ & \textbf{78.9}\textcolor[gray]{0.5}{\small{$\pm1.7$}} & \textbf{86.4}\textcolor[gray]{0.5}{\small{$\pm1.3$}} \\
%     $\checkmark$ & $\checkmark$ & $\times$& 77.6\textcolor[gray]{0.5}{\small{$\pm1.5$}}       & 85.3\textcolor[gray]{0.5}{\small{$\pm3.3$}} \\
%     $\checkmark$ & $\times$& $\checkmark$ & 77.1\textcolor[gray]{0.5}{\small{$\pm3.1$}}       & 84.9\textcolor[gray]{0.5}{\small{$\pm1.5$}} \\
%      $\checkmark$ & $\times$& $\times$ & 73.4\textcolor[gray]{0.5}{\small{$\pm2.7$}}       & 82.1\textcolor[gray]{0.5}{\small{$\pm2.4$}} \\
%       $\times$ & $\times$& $\times$ & 35.1\textcolor[gray]{0.5}{\small{$\pm1.9$}}       & 50.5\textcolor[gray]{0.5}{\small{$\pm1.2$}} \\
%     \bottomrule[1pt]
%     \end{tabular}%
%   \label{tab:ablation}}%
% \end{table}%
\renewcommand{\arraystretch}{1.2}
\begin{table}[t]
  \centering
  \setlength{\tabcolsep}{4pt}
  \caption{Ablation study of EnergyAction components.}
  \resizebox{0.95\linewidth}{!}{
    \begin{tabular}{ccc|cccc}
    \toprule[1pt]
          \multirow{2}{*}{E-Compose} & \multirow{2}{*}{E-Temporal} & \multirow{2}{*}{E-Spatial}& \multicolumn{4}{c}{Avg. Success} \\
    \cmidrule(lr){4-7}
          &  &  & 5-demo & 10-demo & 20-demo & 100-demo \\
    \midrule
    \rowcolor{blue!5!cyan!10}$\checkmark$ & $\checkmark$ & $\checkmark$ & \textbf{44.6}\textcolor[gray]{0.5}{\small{$\pm1.0$}} & \textbf{63.6}\textcolor[gray]{0.5}{\small{$\pm1.3$}} & \textbf{77.3}\textcolor[gray]{0.5}{\small{$\pm3.5$}} & \textbf{86.4}\textcolor[gray]{0.5}{\small{$\pm2.5$}} \\
    $\checkmark$ & $\checkmark$ & $\times$& 43.1\textcolor[gray]{0.5}{\small{$\pm0.9$}} & 62.1\textcolor[gray]{0.5}{\small{$\pm1.1$}}  & 76.6\textcolor[gray]{0.5}{\small{$\pm1.5$}} & 85.3\textcolor[gray]{0.5}{\small{$\pm3.3$}} \\
    $\checkmark$ & $\times$& $\checkmark$ & 42.9\textcolor[gray]{0.5}{\small{$\pm1.3$}} & 61.9\textcolor[gray]{0.5}{\small{$\pm1.6$}} & 76.1\textcolor[gray]{0.5}{\small{$\pm3.1$}} & 84.9\textcolor[gray]{0.5}{\small{$\pm1.5$}} \\
     $\checkmark$ & $\times$& $\times$ & 41.0\textcolor[gray]{0.5}{\small{$\pm1.9$}} & 60.6\textcolor[gray]{0.5}{\small{$\pm2.3$}} & 73.4\textcolor[gray]{0.5}{\small{$\pm2.7$}} & 82.1\textcolor[gray]{0.5}{\small{$\pm2.4$}} \\
      $\times$ & $\times$& $\times$ & 15.2\textcolor[gray]{0.5}{\small{$\pm3.2$}} & 25.6\textcolor[gray]{0.5}{\small{$\pm4.3$}}  & 35.1\textcolor[gray]{0.5}{\small{$\pm1.9$}} & 50.5\textcolor[gray]{0.5}{\small{$\pm1.2$}} \\
    \bottomrule[1pt]
    \end{tabular}%
  \label{tab:ablation}}%
\end{table}%
\begin{figure}[t]
	\centering  
	\includegraphics[width=0.47\textwidth]{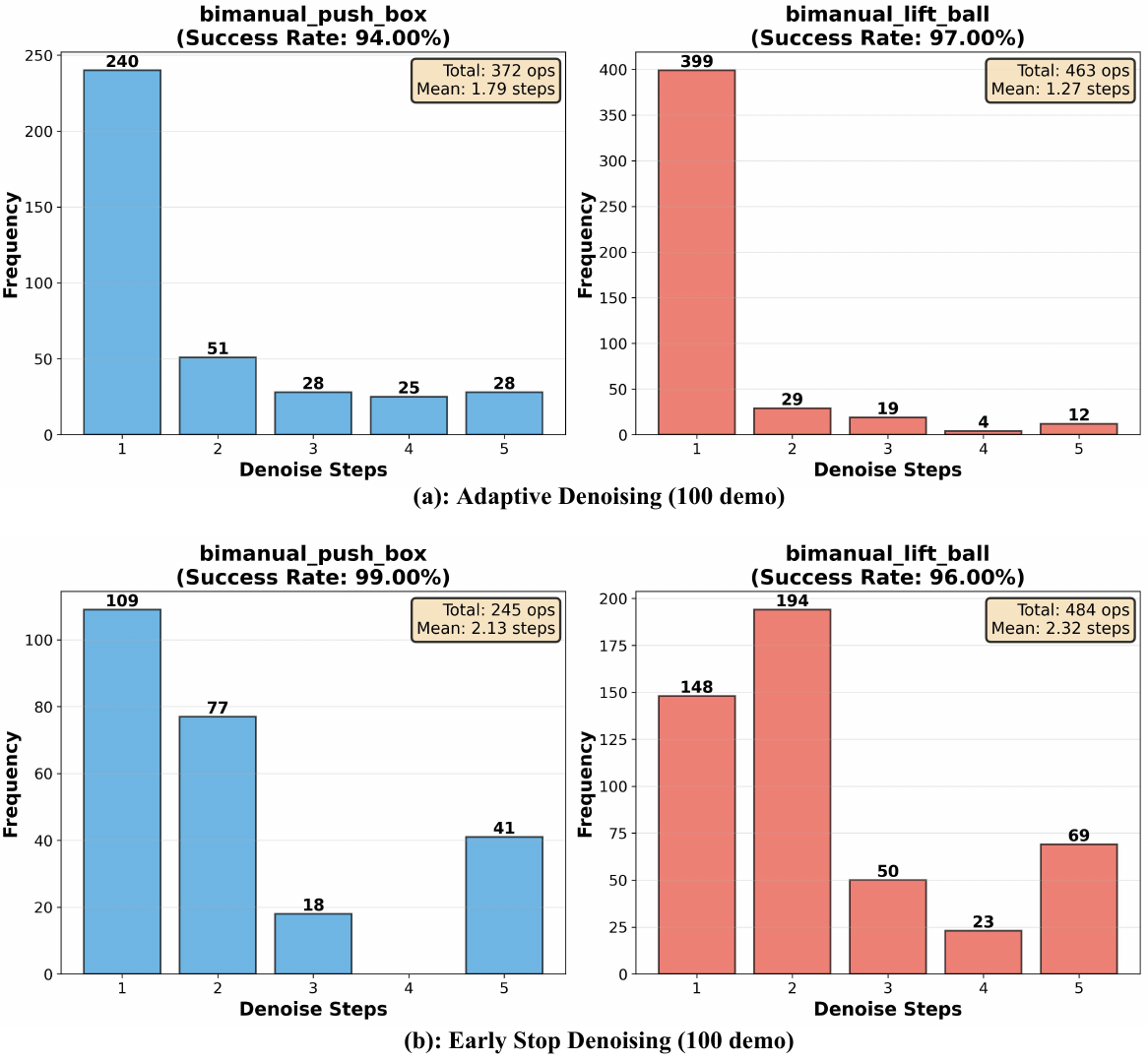}  
	\caption{Distribution of denoising steps for two different energy-aware inference strategies. Both achieve competitive success rates with fast inference speed.}  
	\label{fig:inference} 
\end{figure} 
\textbf{Effects of Model Components.} To systematically evaluate the contribution of each component in EnergyAction, we conduct ablation experiments by progressively removing key modules as shown in Table~\ref{tab:ablation}. The complete model with all three energy components achieves the best performance across both data settings. Removing the spatial energy constraint (E-Spatial) leads to minor performance degradation due to the loss of collision avoidance and spatial feasibility. Similarly, further removing the temporal energy constraint (E-Temporal) results in comparable deterioration, as it enforces crucial action coherence between arms. The impact becomes more pronounced when only the energy-based composition (E-Compose) remains, demonstrating that composition alone without coordination mechanisms is insufficient for effective bimanual manipulation. Most notably, removing all energy-based components yields a dramatic performance collapse, which strongly validates that our energy-based framework is crucial for transferring unimanual knowledge to bimanual tasks. \\
% This progressive degradation pattern highlights the complementary nature of the three components in achieving robust bimanual coordination.
% \renewcommand{\arraystretch}{1.0}
% \begin{table}[t]
%   \centering
%   \caption{Performance comparison of different diffusion policies.}
%   \resizebox{1.0\linewidth}{!}{
%     \begin{tabular}{llccccc}
%     \toprule[1pt]
%     \multirow{2}{*}{Method} & \multirow{2}{*}{Diffusion Policy} & \multicolumn{5}{c}{Avg. Success} \\
%     \cmidrule(lr){3-7}
%     & & 1-demo & 5-demo & 10-demo & 20-demo & 100-demo \\
%     \midrule
%     EnergyAction & DDIM (50 steps) & XX.X\textcolor[gray]{0.5}{\small{$\pm X.X$}} & XX.X\textcolor[gray]{0.5}{\small{$\pm X.X$}} & XX.X\textcolor[gray]{0.5}{\small{$\pm X.X$}} & 18.3\textcolor[gray]{0.5}{\small{$\pm2.0$}} & 24.7\textcolor[gray]{0.5}{\small{$\pm1.8$}} \\
%     EnergyAction & DDPM (100 steps) & XX.X\textcolor[gray]{0.5}{\small{$\pm X.X$}} & XX.X\textcolor[gray]{0.5}{\small{$\pm X.X$}} & XX.X\textcolor[gray]{0.5}{\small{$\pm X.X$}} & 19.1\textcolor[gray]{0.5}{\small{$\pm2.2$}} & 26.2\textcolor[gray]{0.5}{\small{$\pm1.9$}} \\
%     \rowcolor{blue!5!cyan!10}EnergyAction & Flow Matching (5 steps) & XX.X\textcolor[gray]{0.5}{\small{$\pm X.X$}} & XX.X\textcolor[gray]{0.5}{\small{$\pm X.X$}} & XX.X\textcolor[gray]{0.5}{\small{$\pm X.X$}} & \textbf{20.3}\textcolor[gray]{0.5}{\small{$\pm1.5$}} & \textbf{27.0}\textcolor[gray]{0.5}{\small{$\pm1.7$}} \\
%     \bottomrule[1pt]
%     \end{tabular}%
%   \label{tab:diffusion_policies}}%
% \end{table}%
\textbf{Energy-Aware Denoising Strategy Analysis.} 
\begin{figure}[t]
	\centering  
	\includegraphics[width=0.47\textwidth]{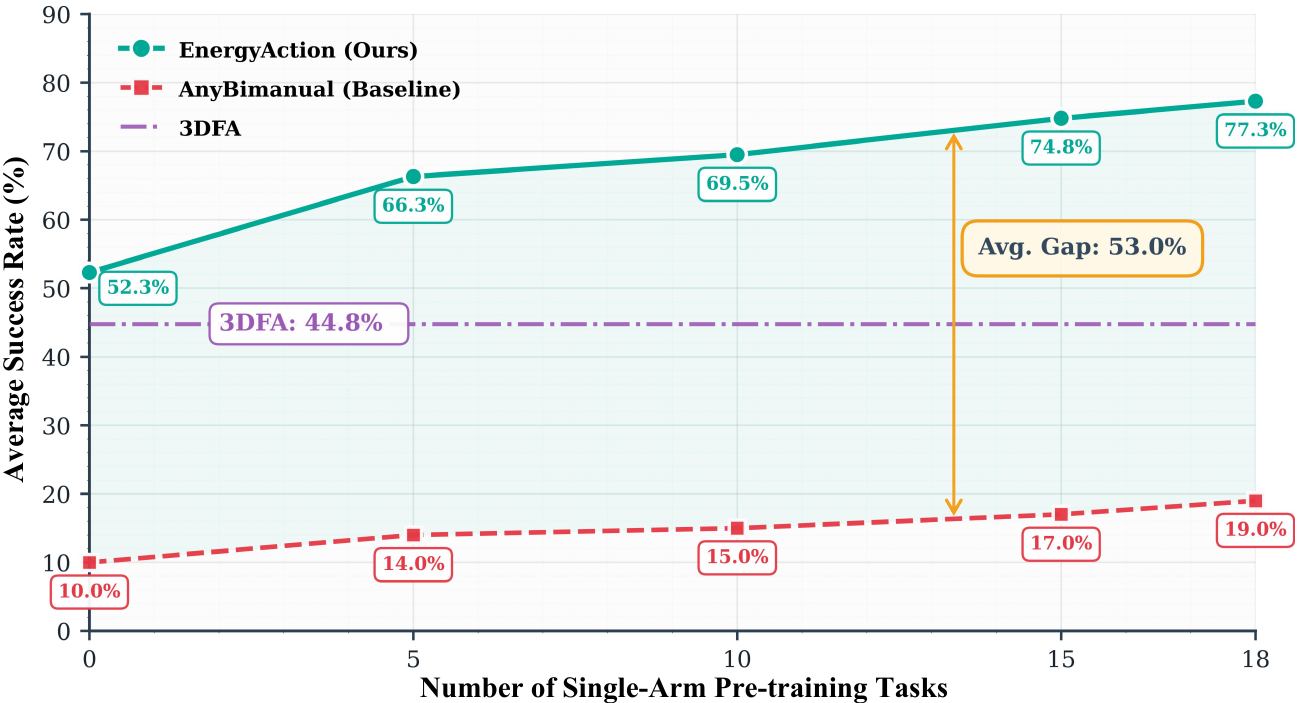}  
	\caption{Single-arm task scaling in EnergyAction.}  
	\label{fig:task_scaling} 
\end{figure} 
\begin{figure*}[t] 
	\centering 
    % \flushleft
	\includegraphics[width=1.0\textwidth]{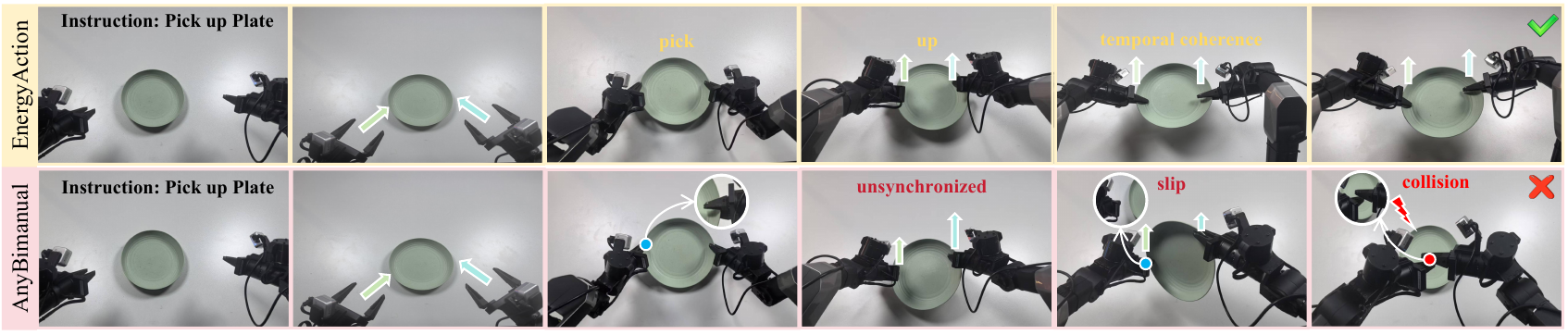} 
        \caption{Visualization of bimanual manipulation in real-world scenarios. {\color[RGB]{146,220,80} Green} and {\color[RGB]{0,176,240} blue} arrows indicate left and right arm motion trajectories. EnergyAction generates coordinated bimanual actions with temporal coherence and spatial feasibility.}
	\label{fig:realtu} 
\end{figure*}
We design two adaptive denoising algorithms based on energy value to accelerate the inference process. As shown in Figure~\ref{fig:inference}, both methods achieve competitive success rates with significantly improved efficiency compared to fixed-step denoising (5 steps). The Adaptive Denoising strategy determines denoising steps based on initial energy values, requiring minimal steps for low-energy actions and maximum steps for high-energy ones. This achieves superior efficiency with mean denoising steps of 1.79 and 1.27. The Early Stop Denoising strategy performs iterative denoising with real-time energy monitoring, terminating once energy falls below a predefined threshold. This ensures quality while maintaining efficiency at mean steps of 2.13 and 2.32. These results demonstrate the advantages of our EnergyAction accelerating inference while maintaining task success rate.\\
% The balanced step distributions indicate that Early Stop adaptively adjusts computation based on actual refinement progress, ensuring reliable action quality with slightly increased computational cost. 
% More detailed analysis and results are provided in Appendix~\ref{sec:More_Ablation}.\\
\textbf{Single-Arm Task Scaling in EnergyAction.}
To investigate how the number of unimanual training tasks affects transfer performance, we conduct experiments under the 20-demonstration setting by varying the number of single-arm tasks from 0 to 18, as shown in Figure~\ref{fig:task_scaling}. First, both EnergyAction and AnyBimanual~\cite{lu2025anybimanual} adopt a similar paradigm of transferring unimanual knowledge to bimanual tasks, and both methods show continuously improving transfer performance as the number of single-arm tasks increases. However, EnergyAction consistently and substantially outperforms AnyBimanual across all task configurations, achieving an average performance gap of 53.0\%. This result demonstrates that our proposed energy composition architecture is more conducive to knowledge transfer from unimanual to bimanual tasks. Most notably, even without any single-arm task pretraining, EnergyAction achieves a 52.3\% success rate, surpassing the 44.8\% of 3DFA~\cite{gkanatsios20253d} that is specifically designed for bimanual tasks. This result provides strong evidence for the effectiveness of our energy-based bimanual policy framework.\\
\textbf{Different Diffusion Policies in EnergyAction.}
\renewcommand{\arraystretch}{1.2}
\begin{table}[t]
  \centering
  \caption{Performance comparison of different unimanual policies.}
  \resizebox{1.0\linewidth}{!}{
    \begin{tabular}{ccccccc}
    \toprule[1pt]
    \multicolumn{2}{c}{Unimanual Policies} & \multicolumn{4}{c}{Avg. Success} \\
    \cmidrule(lr){1-2} \cmidrule(lr){3-6}
    Left-Arm & Right-Arm & 5-demo & 10-demo & 20-demo & 100-demo \\
    \midrule
    DDPM & DDPM & 43.5\textcolor[gray]{0.5}{\small{$\pm 4.7$}} & 63.3\textcolor[gray]{0.5}{\small{$\pm 1.1$}} & 76.2\textcolor[gray]{0.5}{\small{$\pm2.2$}} & 86.1\textcolor[gray]{0.5}{\small{$\pm1.9$}} \\
    DDIM & DDIM & 40.9\textcolor[gray]{0.5}{\small{$\pm 5.2$}} & 60.2\textcolor[gray]{0.5}{\small{$\pm 3.4$}} & 75.3\textcolor[gray]{0.5}{\small{$\pm2.0$}} & 82.1\textcolor[gray]{0.5}{\small{$\pm1.8$}} \\
     DDPM & DDIM & 42.1\textcolor[gray]{0.5}{\small{$\pm3.9$}} & 61.8\textcolor[gray]{0.5}{\small{$\pm2.6$}} & 74.5\textcolor[gray]{0.5}{\small{$\pm2.8$}} & 83.7\textcolor[gray]{0.5}{\small{$\pm2.1$}} \\
    Flow Matching & DDPM & 44.0\textcolor[gray]{0.5}{\small{$\pm2.2$}} & 63.1\textcolor[gray]{0.5}{\small{$\pm1.5$}} & 76.8\textcolor[gray]{0.5}{\small{$\pm3.0$}} & 85.9\textcolor[gray]{0.5}{\small{$\pm2.3$}} \\
    Flow Matching & DDIM & 43.2\textcolor[gray]{0.5}{\small{$\pm2.8$}} & 62.4\textcolor[gray]{0.5}{\small{$\pm1.9$}} & 76.0\textcolor[gray]{0.5}{\small{$\pm2.7$}} & 84.3\textcolor[gray]{0.5}{\small{$\pm2.0$}} \\
    \rowcolor{blue!5!cyan!10}Flow Matching & Flow Matching & \textbf{44.6}\textcolor[gray]{0.5}{\small{$\pm1.0$}} & \textbf{63.6}\textcolor[gray]{0.5}{\small{$\pm1.3$}} & \textbf{77.3}\textcolor[gray]{0.5}{\small{$\pm3.5$}} & \textbf{86.4}\textcolor[gray]{0.5}{\small{$\pm2.5$}} \\
    \bottomrule[1pt]
    \end{tabular}%
  \label{tab:diffusion_policies}}%
\end{table}
To validate the flexibility of our EnergyAction, we perform energy modeling on different unimanual policies and compose them into bimanual policies, including DDPM~\cite{ho2020denoising} (100 denoising steps), DDIM~\cite{song2020denoising} (50 denoising steps), and Flow Matching~\cite{zhang2024robot} (5 denoising steps). As shown in Table~\ref{tab:diffusion_policies}, all model variants achieve competitive performance across different demonstration settings. These consistent results across diverse unimanual policies confirm that our energy-based compositional framework is flexible and robust, with strong compositional reasoning capability that is decoupled from the choice of unimanual policy. 
% , , demonstrating that the framework’s compositional reasoning capability is agnostic to the choice of unimanual policy.
\subsection{Real-world Experiments}
To validate the effectiveness of our approach in real-world scenarios, we conduct experiments following a similar unimanual-to-bimanual transfer paradigm as in simulation. We first train our model on ten single-arm manipulation tasks, with 50 demonstrations per task, then transfer unimanual policies to bimanual coordination tasks. For bimanual tasks, we evaluate the model under two data regimes: 5 demonstrations and 20 demonstrations per task. We test 20 times for each bimanual task to calculate the average success rate. 
As illustrated in Figure~\ref{fig:realtu}, EnergyAction successfully completes the bimanual manipulation task with temporal coherence and spatial feasibility. The left arm (green annotations) and right arm (blue annotations) maintain smooth coordination while satisfying physical constraints. In contrast, AnyBimanual~\cite{lu2025anybimanual} exhibits unsynchronized movements that result in arm collisions, demonstrating its failure to generate coordinated and feasible bimanual actions.
Table~\ref{tab:real_world} presents quantitative comparisons against AnyBimanual~\cite{lu2025anybimanual}, $\pi_0$-keypose~\cite{black2410pi0} and 3DFA~\cite{gkanatsios20253d}. The results demonstrate that EnergyAction consistently outperforms other methods across both data regimes, confirming that our energy-based compositional framework effectively transfers unimanual policies knowledge to bimanual tasks in real-world settings. 
% More experiment details analysis and results are provided in Appendix~\ref{sec:real_world}.

\section{Conclusion}
\renewcommand{\arraystretch}{1.25}
\setlength{\tabcolsep}{3pt}
\begin{table}[t]
  \centering
  \caption{Real-world bimanual tasks performance comparison.}
  \resizebox{\linewidth}{!}{
    \begin{tabular}{ccccc|cc}
    \toprule[1pt]
    \multirow{2}{*}{Method} & \multicolumn{2}{c}{Handover} & \multicolumn{2}{c|}{Pick up Plate} & \multicolumn{2}{c}{Average} \\
    \cmidrule(lr){2-3} \cmidrule(lr){4-5} \cmidrule(lr){6-7}
     & 5-demo & 20-demo & 5-demo & 20-demo & 5-demo & 20-demo \\
    \midrule
    AnyBimanual~\cite{lu2025anybimanual} & 10.0 & 20.0 & 15.0 & 25.0 & 12.5 & 22.5 \\
         % AnyBimanual~\cite{lu2025anybimanual} & 11.0 & 17.0 & 16.0 & 19.0 & 13.5 & 18.0 \\
         $\pi_0$-keypose~\cite{black2410pi0} & 15.0 & 25.0 & 20.0 & 30.0 & 17.5 & 27.5 \\
         3DFA~\cite{gkanatsios20253d} & 20.0 & 30.0 & 25.0 & 40.0 & 22.5 & 35.0 \\
         \rowcolor{blue!5!cyan!10}
    EnergyAction (Ours) & \textbf{35.0} & \textbf{50.0} & \textbf{45.0} & \textbf{55.0} & \textbf{40.0} & \textbf{52.5} \\
    \bottomrule[1pt]
    \end{tabular}
  }
  \label{tab:real_world}
\end{table}

% \renewcommand{\arraystretch}{1.0}
% \setlength{\tabcolsep}{3pt}
% \begin{table}[t]
%   \centering
%   \caption{Real-world bimanual tasks performance comparison.}
%   \resizebox{\linewidth}{!}{
%     \begin{tabular}{lcccc|cc}
%     \toprule[1pt]
%     \multirow{2}{*}{Method} & \multicolumn{2}{c}{Handover} & \multicolumn{2}{c|}{Pick up Plate} & \multicolumn{2}{c}{Average} \\
%     \cmidrule(lr){2-3} \cmidrule(lr){4-5} \cmidrule(lr){6-7}
%      & 5-demo & 20-demo & 5-demo & 20-demo & 5-demo & 20-demo \\
%     \midrule
%     AnyBimanual~\cite{lu2025anybimanual} & 2/20 & 4/20 & 3/20 & 5/20 & 2.5/20 & 4.5/20 \\
%     $\pi_0$-keypose~\cite{black2410pi0} & 3/20 & 5/20 & 4/20 & 6/20 & 3.5/20 & 5.5/20 \\
%     3DFA~\cite{gkanatsios20253d} & 4/20 & 6/20 & 5/20 & 8/20 & 4.5/20 & 7/20 \\
%     \rowcolor{blue!5!cyan!10}
%     EnergyAction & \textbf{7/20} & \textbf{10/20} & \textbf{9/20} & \textbf{11/20} & \textbf{8/20} & \textbf{10.5/20} \\
%     \bottomrule[1pt]
%     \end{tabular}
%   }
%   \label{tab:real_world}
% \end{table}

In this paper, we present EnergyAction, a novel framework that compositionally transfers pre-trained unimanual manipulation policies to bimanual tasks through EBMs. We introduce three key innovations: modeling unimanual policies as composable energy functions, enforcing temporal-spatial coordination via energy-based constraints that ensure coherence and collision avoidance, and developing two different energy-aware denoising strategies that adaptively adjust denoising steps based on action quality assessment for efficient inference. Extensive experiments on simulated and real-world tasks demonstrate that EnergyAction achieves superior performance over existing approaches. We hope this work provides valuable insights for the robotics community and inspires future research in compositional policy learning and energy-based optimization strategies.
\section{Acknowledgements}
We would like to thank all co-authors for their efforts and the reviewers for their constructive comments. This work is supported by the National Natural Science Foundation of China (Key Project of Joint Fund) (Grant No. U24A20328), the National Natural Science Foundation of China (Grant No. 62476071), the Guangdong Basic and Applied Basic Research Foundation (Grant No. 2025A1515011732), the Beijing Natural Science Foundation (Grant Nos. 4262074 and L254018), the National Natural Science Foundation of China (Grant No. 62406092), the National Natural Science Foundation of China (Grant No. U24B20175), the Shenzhen Science and Technology Program (Grant No. KJZD20240903100017022), the Guangdong Basic and Applied Basic Research Foundation (Grant No. 2025A1515010169), and the Shenzhen Science and Technology Program (Grant No. KQTD20240729102207002).
% \clearpage
{
    \small
    \bibliographystyle{ieeenat_fullname}
    \bibliography{energyaction}

@String(CVPR= {IEEE Conf. Comput. Vis. Pattern Recog.})

@String(ICCV= {Int. Conf. Comput. Vis.})

@String(AAAI = {AAAI})

@String(CVPR  = {CVPR})

@String(ICCV  = {ICCV})

@article{xu2024energy,
  title={Energy-based diffusion language models for text generation},
  author={Xu, Minkai and Geffner, Tomas and Kreis, Karsten and Nie, Weili and Xu, Yilun and Leskovec, Jure and Ermon, Stefano and Vahdat, Arash},
  journal={arXiv preprint arXiv:2410.21357},
  year={2024}
}

@article{du2020compositional,
  title={Compositional visual generation with energy based models},
  author={Du, Yilun and Li, Shuang and Mordatch, Igor},
  journal={NeurIPS},
  volume={33},
  pages={6637--6647},
  year={2020}
}

@inproceedings{du2023reduce,
  title={Reduce, reuse, recycle: Compositional generation with energy-based diffusion models and mcmc},
  author={Du, Yilun and Durkan, Conor and Strudel, Robin and Tenenbaum, Joshua B and Dieleman, Sander and Fergus, Rob and Sohl-Dickstein, Jascha and Doucet, Arnaud and Grathwohl, Will Sussman},
  booktitle={ICML},
  pages={8489--8510},
  year={2023},
  organization={PMLR}
}

@inproceedings{lu2025anybimanual,
  title={Anybimanual: Transferring unimanual policy for general bimanual manipulation},
  author={Lu, Guanxing and Yu, Tengbo and Deng, Haoyuan and Chen, Season Si and Tang, Yansong and Wang, Ziwei},
  booktitle={ICCV},
  pages={13662--13672},
  year={2025}
}

@inproceedings{shridhar2023perceiver,
  title={Perceiver-actor: A multi-task transformer for robotic manipulation},
  author={Shridhar, Mohit and Manuelli, Lucas and Fox, Dieter},
  booktitle={CoRL},
  pages={785--799},
  year={2023},
  organization={PMLR}
}

@inproceedings{grotz2024peract2,
  title={Peract2: Benchmarking and learning for robotic bimanual manipulation tasks},
  author={Grotz, Markus and Shridhar, Mohit and Chao, Yu-Wei and Asfour, Tamim and Fox, Dieter},
  booktitle={CoRL 2024 Workshop on Whole-body Control and Bimanual Manipulation: Applications in Humanoids and Beyond},
  year={2024}
}

@article{lecun2006tutorial,
  title={A tutorial on energy-based learning},
  author={LeCun, Yann and Chopra, Sumit and Hadsell, Raia and Ranzato, M and Huang, Fujie and others},
  journal={Predicting structured data},
  volume={1},
  number={0},
  year={2006}
}

@inproceedings{zhang2025energymogen,
  title={Energymogen: Compositional human motion generation with energy-based diffusion model in latent space},
  author={Zhang, Jianrong and Fan, Hehe and Yang, Yi},
  booktitle={CVPR},
  pages={17592--17602},
  year={2025}
}

@article{brohan2022rt,
  title={Rt-1: Robotics transformer for real-world control at scale},
  author={Brohan, Anthony and Brown, Noah and Carbajal, Justice and Chebotar, Yevgen and Dabis, Joseph and Finn, Chelsea and Gopalakrishnan, Keerthana and Hausman, Karol and Herzog, Alex and Hsu, Jasmine and others},
  journal={arXiv preprint arXiv:2212.06817},
  year={2022}
}

@article{gkanatsios20253d,
  title={3D FlowMatch Actor: Unified 3D Policy for Single-and Dual-Arm Manipulation},
  author={Gkanatsios, Nikolaos and Xu, Jiahe and Bronars, Matthew and Mousavian, Arsalan and Ke, Tsung-Wei and Fragkiadaki, Katerina},
  journal={arXiv preprint arXiv:2508.11002},
  year={2025}
}

@article{zhao2023learning,
  title={Learning fine-grained bimanual manipulation with low-cost hardware},
  author={Zhao, Tony Z and Kumar, Vikash and Levine, Sergey and Finn, Chelsea},
  journal={arXiv preprint arXiv:2304.13705},
  year={2023}
}

@inproceedings{goyal2023rvt,
  title={Rvt: Robotic view transformer for 3d object manipulation},
  author={Goyal, Ankit and Xu, Jie and Guo, Yijie and Blukis, Valts and Chao, Yu-Wei and Fox, Dieter},
  booktitle={CoRL},
  pages={694--710},
  year={2023},
  organization={PMLR}
}

@article{ze20243d,
  title={3d diffusion policy: Generalizable visuomotor policy learning via simple 3d representations},
  author={Ze, Yanjie and Zhang, Gu and Zhang, Kangning and Hu, Chenyuan and Wang, Muhan and Xu, Huazhe},
  journal={arXiv preprint arXiv:2403.03954},
  year={2024}
}

@inproceedings{lv2025spatial,
  title={Spatial-temporal graph diffusion policy with kinematic modeling for bimanual robotic manipulation},
  author={Lv, Qi and Li, Hao and Deng, Xiang and Shao, Rui and Li, Yinchuan and Hao, Jianye and Gao, Longxiang and Wang, Michael Yu and Nie, Liqiang},
  booktitle={CVPR},
  pages={17394--17404},
  year={2025}
}

@article{yang2025gripper,
  title={Gripper Keypose and Object Pointflow as Interfaces for Bimanual Robotic Manipulation},
  author={Yang, Yuyin and Cai, Zetao and Tian, Yang and Zeng, Jia and Pang, Jiangmiao},
  journal={arXiv preprint arXiv:2504.17784},
  year={2025}
}

@article{black2410pi0,
  title={$\pi$0: A vision-language-action flow model for general robot control. CoRR, abs/2410.24164, 2024. doi: 10.48550},
  author={Black, Kevin and Brown, Noah and Driess, Danny and Esmail, Adnan and Equi, Michael and Finn, Chelsea and Fusai, Niccolo and Groom, Lachy and Hausman, Karol and Ichter, Brian and others},
  journal={arXiv preprint ARXIV.2410.24164},
year={2024}
}

@article{hu2023towards,
  title={Towards human-robot collaborative surgery: Trajectory and strategy learning in bimanual peg transfer},
  author={Hu, Zhaoyang Jacopo and Wang, Ziwei and Huang, Yanpei and Sena, Aran and y Baena, Ferdinando Rodriguez and Burdet, Etienne},
  journal={IEEE Robotics and Automation Letters},
  volume={8},
  number={8},
  pages={4553--4560},
  year={2023},
  publisher={IEEE}
}

@article{buhl2019dual,
  title={A dual-arm collaborative robot system for the smart factories of the future},
  author={Buhl, Jens F and Gr{\o}nh{\o}j, Rune and J{\o}rgensen, Jan K and Mateus, Guilherme and Pinto, Daniela and S{\o}rensen, Jacob K and B{\o}gh, Simon and Chrysostomou, Dimitrios},
  journal={Procedia manufacturing},
  volume={38},
  pages={333--340},
  year={2019},
  publisher={Elsevier}
}

@article{hinton2002training,
  title={Training products of experts by minimizing contrastive divergence},
  author={Hinton, Geoffrey E},
  journal={Neural computation},
  volume={14},
  number={8},
  pages={1771--1800},
  year={2002},
  publisher={MIT Press}
}

@article{team2024octo,
  title={Octo: An open-source generalist robot policy},
  author={Team, Octo Model and Ghosh, Dibya and Walke, Homer and Pertsch, Karl and Black, Kevin and Mees, Oier and Dasari, Sudeep and Hejna, Joey and Kreiman, Tobias and Xu, Charles and others},
  journal={arXiv preprint arXiv:2405.12213},
  year={2024}
}

@article{liu2024rdt,
  title={Rdt-1b: a diffusion foundation model for bimanual manipulation},
  author={Liu, Songming and Wu, Lingxuan and Li, Bangguo and Tan, Hengkai and Chen, Huayu and Wang, Zhengyi and Xu, Ke and Su, Hang and Zhu, Jun},
  journal={arXiv preprint arXiv:2410.07864},
  year={2024}
}

@inproceedings{varley2024embodied,
  title={Embodied ai with two arms: Zero-shot learning, safety and modularity},
  author={Varley, Jake and Singh, Sumeet and Jain, Deepali and Choromanski, Krzysztof and Zeng, Andy and Chowdhury, Somnath Basu Roy and Dubey, Avinava and Sindhwani, Vikas},
  booktitle={IROS},
  pages={3651--3657},
  year={2024},
  organization={IEEE}
}

@article{darvish2023teleoperation,
  title={Teleoperation of humanoid robots: A survey},
  author={Darvish, Kourosh and Penco, Luigi and Ramos, Joao and Cisneros, Rafael and Pratt, Jerry and Yoshida, Eiichi and Ivaldi, Serena and Pucci, Daniele},
  journal={IEEE Transactions on Robotics},
  volume={39},
  number={3},
  pages={1706--1727},
  year={2023},
  publisher={IEEE}
}

@article{jiang2025rethinking,
  title={Rethinking Bimanual Robotic Manipulation: Learning with Decoupled Interaction Framework},
  author={Jiang, Jian-Jian and Wu, Xiao-Ming and He, Yi-Xiang and Zeng, Ling-An and Wei, Yi-Lin and Zhang, Dandan and Zheng, Wei-Shi},
  journal={arXiv preprint arXiv:2503.09186},
  year={2025}
}

@article{wen2024diffusion,
  title={Diffusion-VLA: Scaling Robot Foundation Models via Unified Diffusion and Autoregression},
  author={Wen, Junjie and Zhu, Minjie and Zhu, Yichen and Tang, Zhibin and Li, Jinming and Zhou, Zhongyi and Li, Chengmeng and Liu, Xiaoyu and Peng, Yaxin and Shen, Chaomin and others},
  journal={arXiv preprint arXiv:2412.03293},
  year={2024}
}

@inproceedings{gao2024bi,
  title={Bi-kvil: Keypoints-based visual imitation learning of bimanual manipulation tasks},
  author={Gao, Jianfeng and Jin, Xiaoshu and Krebs, Franziska and Jaquier, No{\'e}mie and Asfour, Tamim},
  booktitle={ICRA},
  pages={16850--16857},
  year={2024},
  organization={IEEE}
}

@article{zhao2024inverse,
  title={Inverse kinematics solution and control method of 6-degree-of-freedom manipulator based on deep reinforcement learning},
  author={Zhao, Chengyi and Wei, Yimin and Xiao, Junfeng and Sun, Yong and Zhang, Dongxing and Guo, Qiuquan and Yang, Jun},
  journal={Scientific Reports},
  volume={14},
  number={1},
  pages={12467},
  year={2024},
  publisher={Nature Publishing Group UK London}
}

@inproceedings{zhang2025flowpolicy,
  title={Flowpolicy: Enabling fast and robust 3d flow-based policy via consistency flow matching for robot manipulation},
  author={Zhang, Qinglun and Liu, Zhen and Fan, Haoqiang and Liu, Guanghui and Zeng, Bing and Liu, Shuaicheng},
  booktitle={AAAI},
  volume={39},
  number={14},
  pages={14754--14762},
  year={2025}
}

@article{ke20243d,
  title={3d diffuser actor: Policy diffusion with 3d scene representations},
  author={Ke, Tsung-Wei and Gkanatsios, Nikolaos and Fragkiadaki, Katerina},
  journal={arXiv preprint arXiv:2402.10885},
  year={2024}
}

@article{zhao2024aloha,
  title={Aloha unleashed: A simple recipe for robot dexterity},
  author={Zhao, Tony Z and Tompson, Jonathan and Driess, Danny and Florence, Pete and Ghasemipour, Kamyar and Finn, Chelsea and Wahid, Ayzaan},
  journal={arXiv preprint arXiv:2410.13126},
  year={2024}
}

@article{gkanatsios2023energy,
  title={Energy-based models are zero-shot planners for compositional scene rearrangement},
  author={Gkanatsios, Nikolaos and Jain, Ayush and Xian, Zhou and Zhang, Yunchu and Atkeson, Christopher and Fragkiadaki, Katerina},
  journal={arXiv preprint arXiv:2304.14391},
  year={2023}
}

@inproceedings{zitkovich2023rt,
  title={Rt-2: Vision-language-action models transfer web knowledge to robotic control},
  author={Zitkovich, Brianna and Yu, Tianhe and Xu, Sichun and Xu, Peng and Xiao, Ted and Xia, Fei and Wu, Jialin and Wohlhart, Paul and Welker, Stefan and Wahid, Ayzaan and others},
  booktitle={CoRL},
  pages={2165--2183},
  year={2023},
  organization={PMLR}
}

@article{mete2407quest,
  title={Quest: Self-supervised skill abstractions for learning continuous control, 2024},
  author={Mete, Atharva and Xue, Haotian and Wilcox, Albert and Chen, Yongxin and Garg, Animesh},
  journal={URL https://arxiv. org/abs/2407.15840},
  year={2024}
}

@article{ye2024latent,
  title={Latent action pretraining from videos},
  author={Ye, Seonghyeon and Jang, Joel and Jeon, Byeongguk and Joo, Sejune and Yang, Jianwei and Peng, Baolin and Mandlekar, Ajay and Tan, Reuben and Chao, Yu-Wei and Lin, Bill Yuchen and others},
  journal={arXiv preprint arXiv:2410.11758},
  year={2024}
}

@article{yang2024spatiotemporal,
  title={Spatiotemporal predictive pre-training for robotic motor control},
  author={Yang, Jiange and Liu, Bei and Fu, Jianlong and Pan, Bocheng and Wu, Gangshan and Wang, Limin},
  journal={arXiv preprint arXiv:2403.05304},
  year={2024}
}

@article{wang2024dexcap,
  title={Dexcap: Scalable and portable mocap data collection system for dexterous manipulation},
  author={Wang, Chen and Shi, Haochen and Wang, Weizhuo and Zhang, Ruohan and Fei-Fei, Li and Liu, C Karen},
  journal={arXiv preprint arXiv:2403.07788},
  year={2024}
}

@inproceedings{chen2025learning,
  title={Learning coordinated bimanual manipulation policies using state diffusion and inverse dynamics models},
  author={Chen, Haonan and Xu, Jiaming and Sheng, Lily and Ji, Tianchen and Liu, Shuijing and Li, Yunzhu and Driggs-Campbell, Katherine},
  booktitle={ICRA},
  pages={5644--5651},
  year={2025},
  organization={IEEE}
}

@article{sun2024real,
  title={Real-time coordination of multiple robotic arms with reactive trajectory modulation},
  author={Sun, Da and Liao, Qianfang},
  journal={IEEE Transactions on Robotics},
  year={2024},
  publisher={IEEE}
}

@article{zhao2025dual,
  title={A dual-agent collaboration framework based on llms for nursing robots to perform bimanual coordination tasks},
  author={Zhao, Zhendong and Yue, Xiaotian and Xie, Jiexin and Fang, Chuanhong and Shao, Zhenzhou and Guo, Shijie},
  journal={IEEE Robotics and Automation Letters},
  year={2025},
  publisher={IEEE}
}

@article{ho2022classifier,
  title={Classifier-free diffusion guidance},
  author={Ho, Jonathan and Salimans, Tim},
  journal={arXiv preprint arXiv:2207.12598},
  year={2022}
}

@inproceedings{stoop2024method,
  title={A Method for Multi-Robot Asynchronous Trajectory Execution in MoveIt2},
  author={Stoop, Pascal and Ratnayake, Tharaka and Toffetti, Giovanni},
  booktitle={ICRA},
  pages={17694--17700},
  year={2024},
  organization={IEEE}
}

@inproceedings{zhang2024robot,
  title={Robot manipulation with flow matching},
  author={Zhang, Fan and Gienger, Michael},
  booktitle={CoRL 2024 Workshop on Mastering Robot Manipulation in a World of Abundant Data},
  year={2024}
}

@article{zhang2022matching,
  title={Matching in multi-arm bandit with collision},
  author={Zhang, Yirui and Wang, Siwei and Fang, Zhixuan},
  journal={NeurIPS},
  volume={35},
  pages={9552--9563},
  year={2022}
}

@article{song2020denoising,
  title={Denoising diffusion implicit models},
  author={Song, Jiaming and Meng, Chenlin and Ermon, Stefano},
  journal={arXiv preprint arXiv:2010.02502},
  year={2020}
}

@article{ho2020denoising,
  title={Denoising diffusion probabilistic models},
  author={Ho, Jonathan and Jain, Ajay and Abbeel, Pieter},
  journal={NeurIPS},
  volume={33},
  pages={6840--6851},
  year={2020}
}

@book{descartes1912discourse,
  title={A discourse on method},
  author={Descartes, Ren{\'e}},
  year={1912},
  publisher={JM Dent \& Sons Limited}
}

@inproceedings{li2025learning,
  title={Learning precise affordances from egocentric videos for robotic manipulation},
  author={Li, Gen and Tsagkas, Nikolaos and Song, Jifei and Mon-Williams, Ruaridh and Vijayakumar, Sethu and Shao, Kun and Sevilla-Lara, Laura},
  booktitle={ICCV},
  pages={10581--10591},
  year={2025}
}

@inproceedings{kou2025roboannotatorx,
  title={RoboAnnotatorX: A Comprehensive and Universal Annotation Framework for Accurate Understanding of Long-horizon Robot Demonstration},
  author={Kou, Longxin and Ni, Fei and Zheng, Yan and Han, Peilong and Liu, Jinyi and Cui, Haiqin and Liu, Rui and Hao, Jianye},
  booktitle={ICCV},
  pages={10353--10363},
  year={2025}
}

@inproceedings{zhang2025vlabench,
  title={Vlabench: A large-scale benchmark for language-conditioned robotics manipulation with long-horizon reasoning tasks},
  author={Zhang, Shiduo and Xu, Zhe and Liu, Peiju and Yu, Xiaopeng and Li, Yuan and Gao, Qinghui and Fei, Zhaoye and Yin, Zhangyue and Wu, Zuxuan and Jiang, Yu-Gang and others},
  booktitle={ICCV},
  pages={11142--11152},
  year={2025}
}

@inproceedings{yan2025robotron,
  title={RoboTron-Mani: All-in-One Multimodal Large Model for Robotic Manipulation},
  author={Yan, Feng and Liu, Fanfan and Huang, Yiyang and Guan, Zechao and Zheng, Liming and Zhong, Yufeng and Feng, Chengjian and Ma, Lin},
  booktitle={ICCV},
  pages={13707--13718},
  year={2025}
}

@inproceedings{li2025sd2actor,
  title={SD2Actor: Continuous State Decomposition via Diffusion Embeddings for Robotic Manipulation},
  author={Li, Jiayi},
  booktitle={ICCV},
  pages={13751--13760},
  year={2025}
}

@inproceedings{cao2024smart,
  title={Smart help: Strategic opponent modeling for proactive and adaptive robot assistance in households},
  author={Cao, Zhihao and Wang, Zidong and Xie, Siwen and Liu, Anji and Fan, Lifeng},
  booktitle={CVPR},
  pages={18091--18101},
  year={2024}
}

@inproceedings{ryu2024diffusion,
  title={Diffusion-edfs: Bi-equivariant denoising generative modeling on se (3) for visual robotic manipulation},
  author={Ryu, Hyunwoo and Kim, Jiwoo and An, Hyunseok and Chang, Junwoo and Seo, Joohwan and Kim, Taehan and Kim, Yubin and Hwang, Chaewon and Choi, Jongeun and Horowitz, Roberto},
  booktitle={CVPR},
  pages={18007--18018},
  year={2024}
}

@inproceedings{ma2024hierarchical,
  title={Hierarchical diffusion policy for kinematics-aware multi-task robotic manipulation},
  author={Ma, Xiao and Patidar, Sumit and Haughton, Iain and James, Stephen},
  booktitle={CVPR},
  pages={18081--18090},
  year={2024}
}

@inproceedings{zhong2025dexgrasp,
  title={Dexgrasp anything: Towards universal robotic dexterous grasping with physics awareness},
  author={Zhong, Yiming and Jiang, Qi and Yu, Jingyi and Ma, Yuexin},
  booktitle={CVPR},
  pages={22584--22594},
  year={2025}
}

@inproceedings{jia2025lift3d,
  title={Lift3D Policy: Lifting 2D Foundation Models for Robust 3D Robotic Manipulation},
  author={Jia, Yueru and Liu, Jiaming and Chen, Sixiang and Gu, Chenyang and Wang, Zhilve and Luo, Longzan and Li, Xiaoqi and Wang, Pengwei and Wang, Zhongyuan and Zhang, Renrui and others},
  booktitle={CVPR},
  pages={17347--17358},
  year={2025}
}

@inproceedings{yao2025think,
  title={Think Small, Act Big: Primitive Prompt Learning for Lifelong Robot Manipulation},
  author={Yao, Yuanqi and Liu, Siao and Song, Haoming and Qu, Delin and Chen, Qizhi and Ding, Yan and Zhao, Bin and Wang, Zhigang and Li, Xuelong and Wang, Dong},
  booktitle={CVPR},
  pages={22573--22583},
  year={2025}
}

@inproceedings{pan2025omnimanip,
  title={Omnimanip: Towards general robotic manipulation via object-centric interaction primitives as spatial constraints},
  author={Pan, Mingjie and Zhang, Jiyao and Wu, Tianshu and Zhao, Yinghao and Gao, Wenlong and Dong, Hao},
  booktitle={CVPR},
  pages={17359--17369},
  year={2025}
}

@inproceedings{li2025object,
  title={Object-Centric Prompt-Driven Vision-Language-Action Model for Robotic Manipulation},
  author={Li, Xiaoqi and Xu, Jingyun and Zhang, Mingxu and Liu, Jiaming and Shen, Yan and Ponomarenko, Iaroslav and Xu, Jiahui and Heng, Liang and Huang, Siyuan and Zhang, Shanghang and others},
  booktitle={CVPR},
  pages={27638--27648},
  year={2025}
}

@inproceedings{song2025robospatial,
  title={Robospatial: Teaching spatial understanding to 2d and 3d vision-language models for robotics},
  author={Song, Chan Hee and Blukis, Valts and Tremblay, Jonathan and Tyree, Stephen and Su, Yu and Birchfield, Stan},
  booktitle={CVPR},
  pages={15768--15780},
  year={2025}
}

@inproceedings{su2025robosense,
  title={Robosense: Large-scale dataset and benchmark for egocentric robot perception and navigation in crowded and unstructured environments},
  author={Su, Haisheng and Song, Feixiang and Ma, Cong and Wu, Wei and Yan, Junchi},
  booktitle={CVPR},
  pages={27446--27455},
  year={2025}
}

@inproceedings{wang2025flowram,
  title={FlowRAM: Grounding Flow Matching Policy with Region-Aware Mamba Framework for Robotic Manipulation},
  author={Wang, Sen and Wang, Le and Zhou, Sanping and Tian, Jingyi and Li, Jiayi and Sun, Haowen and Tang, Wei},
  booktitle={CVPR},
  pages={12176--12186},
  year={2025}
}

@inproceedings{mu2025robotwin,
  title={Robotwin: Dual-arm robot benchmark with generative digital twins},
  author={Mu, Yao and Chen, Tianxing and Chen, Zanxin and Peng, Shijia and Lan, Zhiqian and Gao, Zeyu and Liang, Zhixuan and Yu, Qiaojun and Zou, Yude and Xu, Mingkun and others},
  booktitle={CVPR},
  pages={27649--27660},
  year={2025}
}

@inproceedings{zhou2025physvlm,
  title={Physvlm: Enabling visual language models to understand robotic physical reachability},
  author={Zhou, Weijie and Tao, Manli and Zhao, Chaoyang and Guo, Haiyun and Dong, Honghui and Tang, Ming and Wang, Jinqiao},
  booktitle={CVPR},
  pages={6940--6949},
  year={2025}
}

@article{tomaselli2024synchronization,
  title={Synchronization of moving chaotic robots},
  author={Tomaselli, Cinzia and Guastella, Dario C and Muscato, Giovanni and Minati, Ludovico and Frasca, Mattia and Gambuzza, Lucia Valentina},
  journal={IEEE Robotics and Automation Letters},
  volume={9},
  number={7},
  pages={6496--6503},
  year={2024},
  publisher={IEEE}
}

@inproceedings{qi2025two,
  title={Two by two: Learning multi-task pairwise objects assembly for generalizable robot manipulation},
  author={Qi, Yu and Ju, Yuanchen and Wei, Tianming and Chu, Chi and Wong, Lawson LS and Xu, Huazhe},
  booktitle={CVPR},
  pages={17383--17393},
  year={2025}
}

@article{song2025few,
  title={Few-shot vision-language action-incremental policy learning},
  author={Song, Mingchen and Deng, Xiang and Zhong, Guoqiang and Lv, Qi and Wan, Jia and Li, Yinchuan and Hao, Jianye and Guan, Weili},
  journal={arXiv preprint arXiv:2504.15517},
  year={2025}
}

@inproceedings{zhang2024multi,
  title={Multi-factor adaptive vision selection for egocentric video question answering},
  author={Zhang, Haoyu and Liu, Meng and Liu, Zixin and Song, Xuemeng and Wang, Yaowei and Nie, Liqiang},
  booktitle={Forty-first International Conference on Machine Learning},
  year={2024}
}

@article{li2026global,
  title={Global Prior Meets Local Consistency: Dual-Memory Augmented Vision-Language-Action Model for Efficient Robotic Manipulation},
  author={Li, Zaijing and Hu, Bing and Shao, Rui and Chen, Gongwei and Jiang, Dongmei and Xie, Pengwei and Hao, Jianye and Nie, Liqiang},
  journal={arXiv preprint arXiv:2602.20200},
  year={2026}
}
}
\end{document}